\documentclass[11pt]{article}

\usepackage[preprint]{acl}

\usepackage{times}
\usepackage{latexsym}
\usepackage[T1]{fontenc}
\usepackage[utf8]{inputenc}
\usepackage{microtype}
\usepackage{inconsolata}
\usepackage{graphicx}
\usepackage{amsmath}
\usepackage{amssymb}
\usepackage{booktabs}
\usepackage{siunitx}
\usepackage{tabularx}
\usepackage{array}
\usepackage{colortbl}
\usepackage{float}
\usepackage{placeins}
\usepackage{makecell}
\usepackage{algorithm}
\usepackage{algpseudocode}
\usepackage{CJKutf8}
\usepackage[most]{tcolorbox}

\newcolumntype{Y}{>{\raggedright\arraybackslash}X}
\newcommand{\NA}{--}
\newcommand{\robusteq}{\mathrel{\boldsymbol{=}}}
\newcommand{\stageto}[1]{\mathrel{\overset{\mathrm{#1}}{\longrightarrow}}}
\definecolor{promptboxback}{RGB}{244,244,244}
\definecolor{promptboxframe}{RGB}{70,78,92}
\definecolor{promptboxtitle}{RGB}{170,170,170}
\definecolor{exampleboxback}{RGB}{250,250,247}
\definecolor{exampleboxframe}{RGB}{108,108,95}
\definecolor{exampleboxtitle}{RGB}{238,238,229}
\definecolor{tablegroupgray}{RGB}{232,232,232}

\newenvironment{promptbox}[1]{%
  \begin{tcolorbox}[
    enhanced jigsaw,
    breakable,
    colback=promptboxback,
    colframe=promptboxframe,
    colbacktitle=promptboxtitle,
    coltitle=black,
    title={#1},
    fonttitle=\bfseries\footnotesize,
    fontupper=\footnotesize,
    halign title=center,
    boxrule=0.45pt,
    arc=4pt,
    outer arc=4pt,
    boxsep=2pt,
    left=7pt,
    right=7pt,
    top=5pt,
    bottom=5pt,
    before skip=6pt,
    after skip=2pt
  ]%
  \setlength{\parskip}{0.25em}\noindent\raggedright
}{%
  \end{tcolorbox}%
}
\newenvironment{promptfigure}[2]{%
  \def\promptfigurecaption{#2}%
  \begin{promptbox}{#1}%
}{%
  \end{promptbox}%
  {\captionsetup{type=figure,skip=3pt}\captionof{figure}{\promptfigurecaption}}%
  \par\vspace{0.75em}%
}
\newcounter{appendixexample}[section]
\renewcommand{\theappendixexample}{\thesection.\arabic{appendixexample}}

\title{\textsc{\textbf{LegalWorld}}: A Life-Cycle Interactive Environment for Legal Agents}

\author{
\textbf{Songhan Zuo}$^{1,2\dagger}$, \textbf{Shengbin Yue}$^{1\dagger}$, \textbf{Tao Chiang}$^{1}$, \textbf{Guanying Li}$^{1}$, \textbf{Yun Song}$^{3}$,\\
\textbf{Xuanjing Huang}$^{1,2}$, \textbf{Zhongyu Wei}$^{1,2*}$\\
$^{1}$Fudan University \quad $^{2}$Shanghai Innovation Institute\\
$^{3}$Northwest University of Political and Law\\
\texttt{songhanzuo@gmail.com, sbyue23@m.fudan.edu.cn, zywei@fudan.edu.cn}\\
Project Page: \url{https://chidaic.github.io/Legal-world/}
}

\begin{document}
\pagestyle{plain}
\maketitle
\thispagestyle{plain}

\begin{abstract}
    Civil litigation is inherently a life-cycle process: what a lawyer drafts on day one constrains what unfolds at trial months later. Yet existing legal benchmarks evaluate isolated subtasks, and prior legal-agent simulators reinitialize each scenario from shared ground truth, leaving cross-stage causal dependencies unmodeled. We present \textsc{\textbf{LegalWorld}}, a life-cycle interactive environment that models Chinese civil litigation as a causally connected state chain of five stages (seven sub-scenarios), grounded in 75{,}309 paired Chinese civil judgments. We pair it with reusable infrastructure (local memory, global case memory, a Skill/Tool library) that keeps each dispute consistent across its full life cycle. Building on this environment, we construct \textbf{LongJud-Bench} to evaluate agent capability across all five connected stages. 18{,}992 ratings from 217 legal-background evaluators confirm that \textsc{\textbf{LegalWorld}} trajectories are procedurally faithful and role-consistent; and a capability-level cross-model evaluation reveals sharp divergences that aggregate scores cannot expose, with no single backbone leading across consultation, drafting, and courtroom advocacy. Detailed resources will be released publicly.
\end{abstract}

\section{Introduction}

Legal artificial intelligence has made substantial progress in recent years, spanning legal language models \citep{cui2024chatlaw,yue2023disclawllm}, evaluation benchmarks \citep{fei-etal-2024-lawbench,guha2023legalbench,xiao2018cail,li-etal-2024-lexeval}, and interactive agent systems \citep{chen_agentcourt_nodate,he-etal-2024-agentscourt-emnlp,jia2026readyjuristonebenchmarking}. These advances, however, remain largely confined to single-scenario settings, where each task is evaluated over fixed inputs without inheriting state from earlier procedural stages, a limitation echoed by recent calls to reframe legal-agent benchmarking around realistic workflows and agentic performance \citep{ranjan_motivations_nodate,liu_llm_2026}. Real civil litigation, by contrast, is not a collection of independent tasks. A dispute unfolds from initial consultation through document drafting, first-instance trial, appeal, and second-instance judgment, with facts, claims, evidence, and procedural choices from earlier stages shaping what can happen later. Each stage consumes the artifacts produced by the previous stage; drafting errors propagate downstream into trial outcomes; party knowledge, lawyer strategies, and judicial findings co-evolve along a single causal chain. Modeling this complete litigation life cycle is therefore a prerequisite for assessing whether a legal agent possesses genuine procedural capability rather than isolated task skills.

\begin{figure}[!t]
  \centering
  \includegraphics[width=\linewidth]{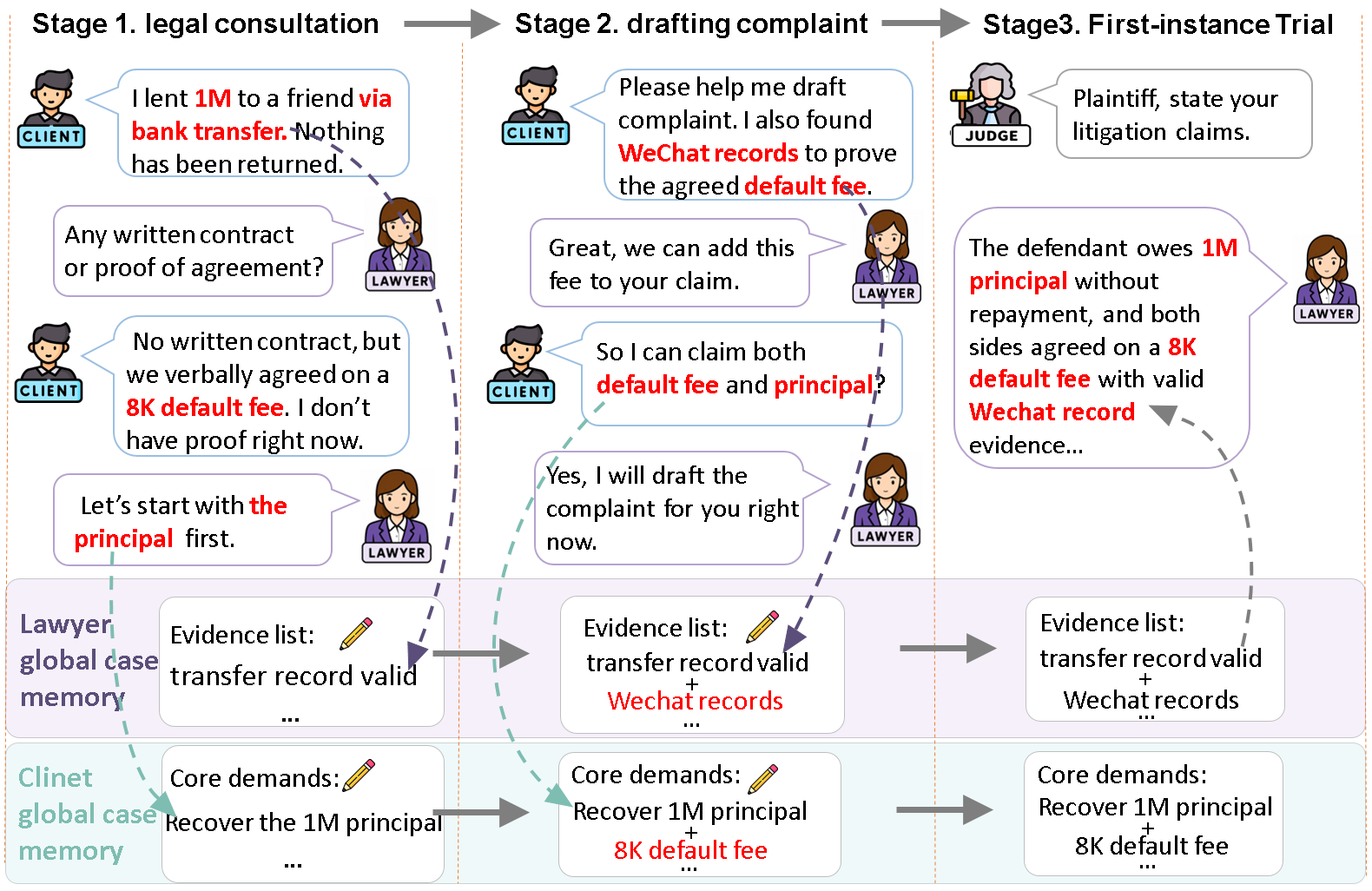}
  \caption{Example from \textsc{\textbf{LegalWorld}}. The figure traces a civil dispute from legal consultation to the first-instance civil trial, showing scene-level communication content and the memory flow through which case information is recorded, updated, and carried forward.}
  \label{fig:consultation-first-instance-example}
\end{figure}

Recent legal-agent simulators take a step toward more realistic legal scenarios, yet three key gaps remain. \textbf{(1) Long-horizon stage coverage.} Existing systems cover only local segments of the process, modeling adversarial courtroom procedures alone \citep{chen_agentcourt_nodate,he-etal-2024-agentscourt-emnlp} or initializing each scenario from shared case-level ground truth rather than from the previous scenario's output \citep{jia2026readyjuristonebenchmarking}, leaving cross-stage state transmission structurally missing. \textbf{(2) Heterogeneous role consistency.} Clients, lawyers, and judges hold distinct knowledge horizons and adversarial stances that continuously evolve as the case proceeds, yet existing simulators reinitialize each scenario from shared ground truth and cannot preserve this stage-bound role state. \textbf{(3) Procedural tool support.} Real legal tasks require dedicated tool and skill support for evidence submission, document drafting, and courtroom procedure, which current agent environments rarely provide. Together, these gaps point to a common requirement: a complete life-cycle simulation environment with role-bound interfaces and procedural infrastructure.

To address the three gaps above, we propose \textsc{\textbf{LegalWorld}}, a life-cycle interactive environment for legal agents. Figure~\ref{fig:consultation-first-instance-example} illustrates a concrete dispute trajectory. \textsc{\textbf{LegalWorld}} models Chinese civil litigation as a five-stage causal chain across seven sub-scenarios, where each stage consumes facts, evidence, positions, and documents from earlier stages, forming a causally connected trajectory over the full life cycle. The environment construction is supported by 75{,}309 paired first- and second-instance Chinese civil cases covering over 500 causes of action. Three agent types---clients, lawyers, and judges---are instantiated through role-specific, stage-bound interfaces with appropriate visibility, actions, and Skill/Tool access. For long-horizon simulation, it provides reusable infrastructure: in-scenario local memory, global case memory, and a modular Skill/Tool library, which together keep facts, evidence, and positions consistent as the case advances through its stages.

Building on this environment, we construct \textbf{LongJud-Bench} to evaluate the life-cycle legal capability of agents across all five connected stages of \textsc{\textbf{LegalWorld}}. A large-scale human study with 18{,}992 ratings from 217 legal-background evaluators confirms that \textsc{\textbf{LegalWorld}} trajectories are procedurally faithful and role-consistent, establishing a reliable testbed for legal-agent research. Cross-model evaluation on LongJud-Bench further reveals capability-level divergences across backbones that aggregate scores cannot expose, with no single backbone leading across consultation, drafting, and courtroom advocacy.

Our contributions are: \textbf{(A) The first life-cycle civil litigation simulation environment.} We construct \textsc{\textbf{LegalWorld}}, which simulates Chinese civil litigation from consultation to final second-instance judgment as a five-stage state chain across seven sub-scenarios; \textbf{(B) Reusable infrastructure for long-horizon legal agents.} We design in-scenario local memory, global case memory, and a modular Skill/Tool library that keep case state consistent across the full litigation life cycle; and \textbf{(C) A life-cycle legal capability benchmark.} Based on \textsc{\textbf{LegalWorld}}, we build LongJud-Bench to evaluate individual legal capability over the full litigation life cycle.



\section{\textsc{\textbf{LegalWorld}}: Constructing a Life-Cycle Civil Litigation Environment}

\textsc{\textbf{LegalWorld}} turns real civil cases into runnable life-cycle litigation trajectories. Starting from paired first- and second-instance judgments, the environment extracts a structured case seed, initializes role and persona conditions, exposes stage-specific procedural interfaces, records agent interaction traces and stage outputs, and updates the case state after each stage.

\begin{figure*}[!t]
  \centering
  \includegraphics[width=\textwidth]{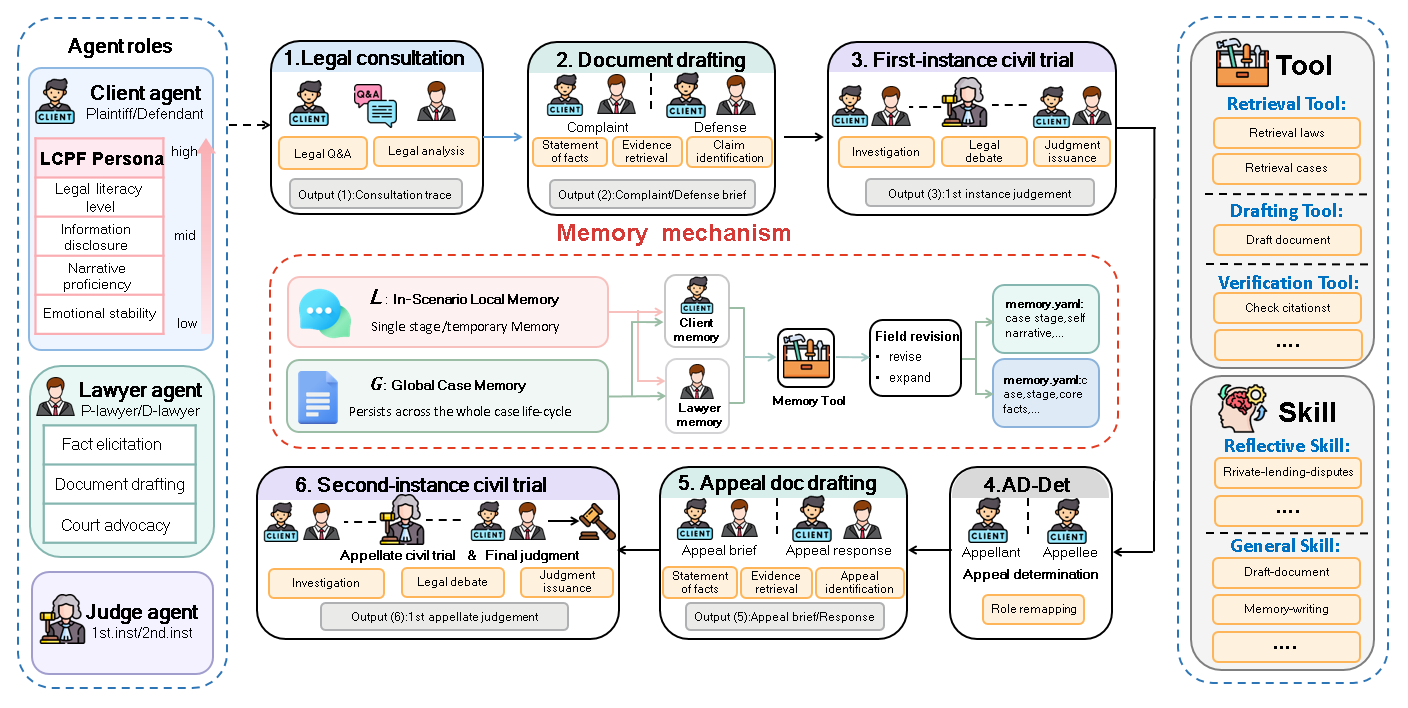}
  \caption{Overview of \textsc{\textbf{LegalWorld}}. The figure shows the participating client, lawyer, and judge agents, the five-stage life-cycle state chain, in-scenario local memory, global case memory, and Skill/Tool support.}
  \label{fig:legal-world-overview}
\end{figure*}

Figure~\ref{fig:legal-world-overview} gives the runtime organization of \textsc{\textbf{LegalWorld}}, including the participating roles, life-cycle stage chain, and the support components connected to the simulation process.

We organize this section into four parts: data-driven case construction (\S\ref{sec:data-foundation}), life-cycle state and interface design (\S\ref{sec:lifecycle-state}), role and persona initialization (\S\ref{sec:role-interface}), and the stage construction protocol (\S\ref{sec:stage-protocol}). Together, these parts turn each case into a connected litigation trajectory that can be instantiated by different LLM backbones under the same procedural interface.

\subsection{Data-Driven Case Construction}
\label{sec:data-foundation}

A life-cycle environment is only as faithful as the cases that drive it, so \textsc{\textbf{LegalWorld}} is grounded in real civil litigation data rather than manually invented disputes. The construction pipeline turns raw public judgments into runnable case seeds in four steps---source collection, first/second-instance pairing, structured field extraction, and persona/consultation generation---and then exposes the resulting fields to agents only through stage-specific visibility rules. Formally, let $\mathcal{D}$ denote the paired judgment collection, with each case $c$ containing a first-instance judgment $J^{(1)}_c$ and a second-instance judgment $J^{(2)}_c$; we convert each pair into a structured case seed $D_c$ that the environment instantiates as a connected litigation trajectory.

\paragraph{Source collection.} We collect public civil first-instance and second-instance judgment documents from China Judgments Online (\textit{wenshu.court.gov.cn}), retaining the judgment text and case number of each document and removing duplicate filings. The resulting corpus spans courts at every level of the Chinese civil court hierarchy and forms the raw material from which runnable cases are built.

\begin{figure}[!t]
  \centering
  \includegraphics[width=\linewidth]{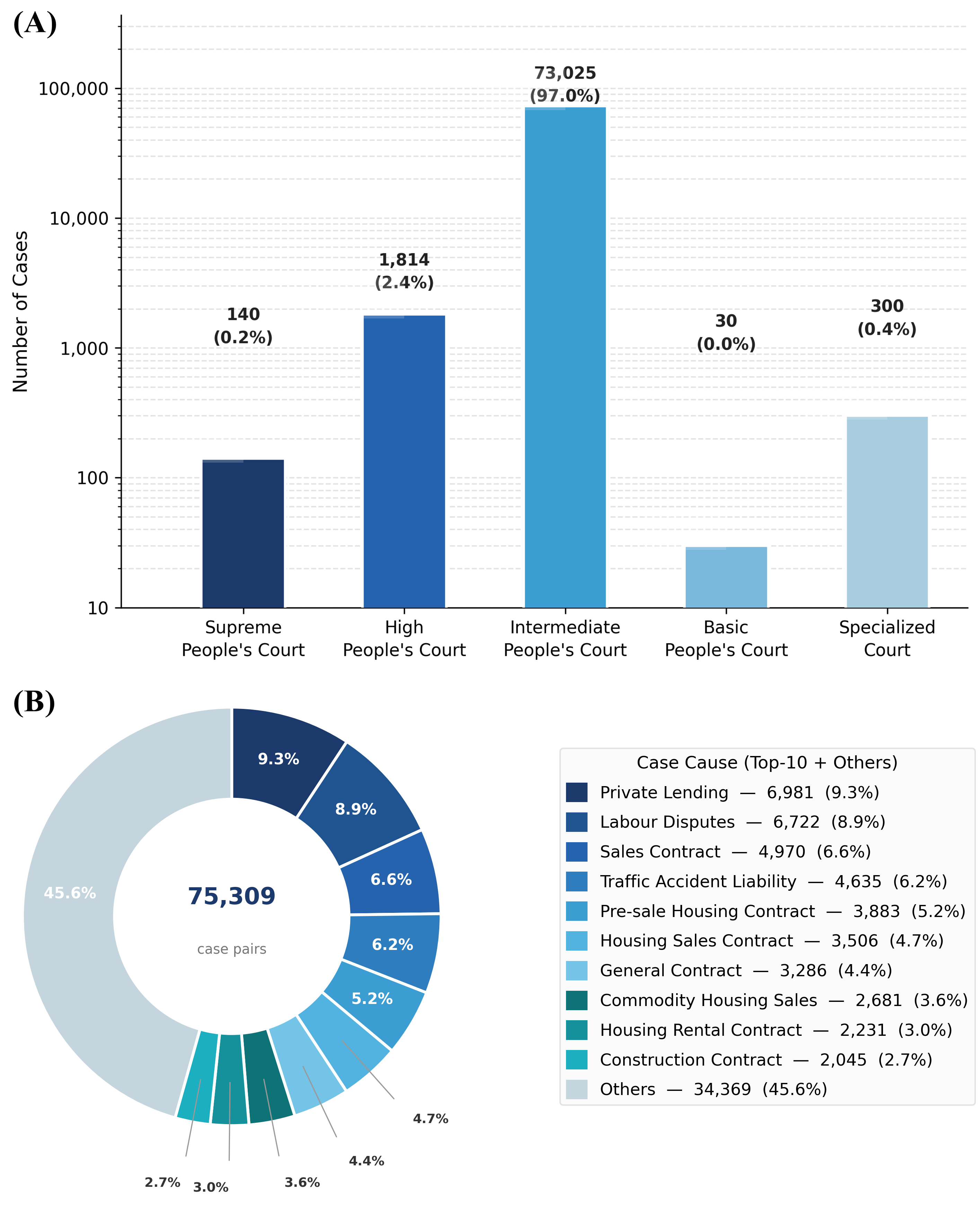}
    \caption{Data foundation for \textsc{\textbf{LegalWorld}} environment construction. (A) Court-level distribution of the 75,309 second-instance judgments used to construct runnable civil-litigation case trajectories. Most cases are decided at the intermediate court level, consistent with the structure of Chinese civil appellate jurisdiction. (B) Top-category cause-of-action distribution across all 75,309 paired cases. The distribution summarizes the broad legal-domain coverage available for environment construction and shows that \textsc{\textbf{LegalWorld}} supports both frequent and long-tail civil disputes.}
  \label{fig:dataset-composition}
\end{figure}

\paragraph{First/second-instance pairing.} We pair the first- and second-instance judgments of the same dispute, matching each case by shared case number, identical parties, and consistent cause of action. We further drop second-instance records that never reached a substantive appellate hearing (e.g., withdrawal, non-acceptance, or procedural dismissal), so that every retained pair carries a genuine first-to-second-instance progression. After pairing and filtering, 75{,}309 (first, second) tuples remain, covering over 500 causes of action and spanning both high-frequency and long-tail civil disputes (Figure~\ref{fig:dataset-composition}).

\paragraph{Structured field extraction.} We convert each paired judgment into a structured case seed $D_c$ that reorganizes the two judgments into the fields a litigation trajectory needs. Using a stage-typed schema, the seed records case metadata, party fields, claims and defenses, facts and reasons, evidence lists, first-instance court findings and disposition, and the analogous appeal and second-instance fields, so that one $D_c$ captures the full procedural record of a dispute. The seed is not exposed to agents as a whole: the environment later releases its fields through the stage-specific visibility rules of Section~\ref{sec:lifecycle-state}, so each agent observes only what its role and stage permit. The extraction model, schema, and quality-control procedure are described in Appendix~\ref{app:dataset}.

\paragraph{Persona and consultation seeds.} Two further generation steps make each seed runnable as an interactive dispute rather than a static record. First, we assign each litigant a persona under the Legal Client Persona Framework (LCPF, Section~\ref{sec:role-interface}), which conditions how the party discloses facts, asks questions, and reacts during the simulation. Second, conditioned on the LCPF persona and the accepted facts, we generate party-side consultation questions together with reference answers grounded in the applicable statutes; the questions drive the consultation stage, while the reference answers are reserved for evaluation only.

\paragraph{Scale and splits.} The complete corpus (\textbf{Full}) retains all 75{,}309 paired cases. To keep large-scale simulation tractable---one complete life-cycle run averages about 500{,}000 tokens---we additionally derive two cause-balanced subsets under a fixed seed: \textbf{Medium} (1{,}000 cases from the top 100 causes) and \textbf{Light} (100 cases from the top 20 causes). Figure~\ref{fig:dataset-composition} summarizes the court-level and top-category cause-of-action distribution of the corpus, and Appendix~\ref{app:dataset} (Table~\ref{tab:dataset-splits}) reports the per-split sizes and sampling rules.

\subsection{Life-Cycle State and Interface Design}
\label{sec:lifecycle-state}

Given a case seed $D_c$, \textsc{\textbf{LegalWorld}} instantiates the same civil dispute as a connected multi-agent trajectory over the life-cycle stages. The participating agent set $\mathcal{A}_c$ contains the plaintiff client $a_p$, defendant client $a_d$, plaintiff lawyer $a_{lp}$, defendant lawyer $a_{ld}$, first-instance judge $a_{j1}$, and second-instance judge $a_{j2}$; $\mathcal{A}_c^{(t)}$ denotes the subset active at stage $t$. The role mapping at stage $t$ is denoted by $R_c^{(t)}$; after the appeal-determination transition, $R_c^{(t)}$ maps the original plaintiff and defendant sides into appellant or appellee roles according to the appeal fields in $D_c$.

The life cycle comprises five connected stages, instantiated through seven concrete sub-scenarios: Legal Consultation (LC), Complaint Drafting (CD), Defense Drafting (DD), First-Instance Trial (FIT), Appeal Drafting (AD), Appeal Response (AR), and Second-Instance Trial (SIT):
\begin{equation}
\begin{aligned}
S_c^{(0)}
  &\stageto{LC} S_c^{(1)}
  \stageto{CD/DD} S_c^{(2)}
  \stageto{FIT} S_c^{(3)} \\
  &\stageto{AD/AR} S_c^{(4)}
  \stageto{SIT} S_c^{(5)} .
\end{aligned}
\label{eq:state-chain}
\end{equation}

The stage state $S_c^{(t)}$ records the case seed, role mapping, accumulated artifacts, interaction traces, and memory handle:
\begin{equation}
S_c^{(t)}\robusteq
\left(D_c,R_c^{(t)},O_c^{(\leq t)},H_c^{(\leq t)},M_c^{(t)}\right).
\label{eq:stage-state}
\end{equation}
For each agent $a$ at stage $t$, the role-specific interface is
\begin{equation}
I_{a,c}^{(t)}\robusteq
\left(V_{a,c}^{(t)},\Phi_a^{(t)},\Sigma_a^{(t)},\mathcal{U}_a^{(t)}\right).
\label{eq:stage-interface}
\end{equation}
Here, $V_{a,c}^{(t)}$ is the role-visible state derived from $S_c^{(t-1)}$ and $D_c$, $\Phi_a^{(t)}$ is the stage procedural template, $\Sigma_a^{(t)}$ is the Skill/Tool support entry from Section~\ref{sec:tool-skill-supply}, and $\mathcal{U}_a^{(t)}$ is the permitted action set.

\subsection{Role and Persona Initialization}

\label{sec:role-interface}

\textsc{\textbf{LegalWorld}} instantiates three agent types---\textbf{lawyers} $\{a_{lp},a_{ld}\}$, \textbf{clients} $\{a_p,a_d\}$, and \textbf{judges} $\{a_{j1},a_{j2}\}$---each constructed from a role profile, stage-specific visibility rules, permitted action types, and a Skill/Tool boundary exposed through $I_{a,c}^{(t)}$. One lawyer serves as the target agent $a^\ast$ under evaluation; the other supplies adversarial counterpart behavior. Clients carry party-side narratives under the persona conditions defined below. Judges are stage-bound, do not write persistent memory, and produce judgment artifacts at FIT and SIT only. Appendix~\ref{app:scenario-state} lists role profiles, visible-state rules, and permitted actions for all three agent types.

\paragraph{Legal Client Persona Framework (LCPF).}
Prior legal-agent and social-agent environments often rely on broad, general-purpose persona traits \citep{jia2026readyjuristonebenchmarking,zhou2024sotopia}. We find these too coarse for ordinary litigants, who differ less in broad personality than in how they understand legal procedure, disclose facts, tolerate procedural pressure, and organize case narratives. Inspired by PatientSim's domain-specific persona design \citep{patientsim2025physionet}, LCPF defines four legal-scene dimensions---Legal Literacy, Information Disclosure Willingness, Emotional Stability, and Narrative Proficiency---each at high, medium, or low. Their combinations shape disclosure, question-asking, risk reaction, and evidence narration in the simulation (Appendix~\ref{app:dataset-construction-details}).

\subsection{Stage Construction Protocol}
\label{sec:stage-protocol}

All five life-cycle stages share one construction pattern: the environment reads $S_c^{(t-1)}$, assigns roles via $R_c^{(t)}$, exposes interface $I_{a,c}^{(t)}$ to each agent, records the dialogue trace $H_c^{(t)}$ and legal artifact $O_c^{(t)}$, and appends them to $S_c^{(t)}$ through a transition function. The concrete scenarios are as follows. \textbf{Legal Consultation (LC)} builds the initial client-lawyer interaction from persona-conditioned facts and party questions, producing consultation records and lawyer advice. At the pre-trial drafting stage, \textbf{Complaint Drafting (CD)} is used when the target lawyer represents the plaintiff, while \textbf{Defense Drafting (DD)} is used for the defendant; both collect party facts, claims or defenses, and evidence, then generate first-instance pleading artifacts. \textbf{First-Instance Trial (FIT)} brings both parties, both lawyers, and the first-instance judge into a structured trial that produces a transcript and first-instance judgment.

After FIT, \textbf{Appeal Determination (AD-Det)} is an environment transition rather than an agent-driven stage: it reads the appeal fields in $D_c$ and remaps the original plaintiff/defendant sides into appellant/appellee roles. The pre-appellate drafting stage then uses \textbf{Appeal Drafting (AD)} for the appellant side or \textbf{Appeal Response (AR)} for the appellee side, generating appellate pleadings and drafting traces from the first-instance judgment, appeal requests, and new evidence when available. \textbf{Second-Instance Trial (SIT)} follows the structured trial procedure with appellate role titles and produces the final judgment $J_{\mathrm{final}}\robusteq O_c^{(5)}$. CD/DD and AD/AR are role-conditional: only the sub-scenario triggered by $a^\ast$'s procedural side is executed and scored. FIT and SIT share the trial procedure in Appendix Algorithm~\ref{alg:structured-trial}, with role titles adapted to the appellate context.

\section{Life-Cycle Environment Infrastructure}

Section~2 defines the life-cycle state chain $S_c^{(t)}$ and the stage interface $I_{a,c}^{(t)}$. To make this chain runnable over a long simulation, \textsc{\textbf{LegalWorld}} provides two runtime components: the memory handle $M_c^{(t)}$ and the Skill/Tool support entry $\Sigma_a^{(t)}$.

\subsection{Life-Cycle Environment Memory Infrastructure}
\label{sec:cross-stage-memory}

Each case in \textsc{\textbf{LegalWorld}} is accompanied by structured memories for participating clients and lawyers \citep{packer2024memgpt,zhang_memskill_2026}. For a memory-maintaining agent $a$, the agent-level memory handle $M_{a,c}^{(t)}$ separates \textbf{in-scenario local memory} $L_{a,c}^{(t)}$ from \textbf{global case memory} $G_{a,c}^{(t)}$. Local memory is the dialogue record exposed back to agents inside a single scenario---the portion of $H_c^{(t)}$ that preserves turn-level continuity. It does not consolidate dialogue into durable facts; that structured consolidation is handled by global memory at stage end. Global case memory stores information that should persist across stages within the same case: facts, evidence status, claims and defenses, procedural progress, client goals, and confirmed litigation positions. The case-level handle $M_c^{(t)}$ aggregates the agent-level handles for $\mathcal{A}_{\mathrm{mem},c}^{(t)}$, the clients and lawyers with memory-writing responsibility. Judge agents do not write persistent memory because they are instantiated as stage-specific roles. At the end of each stage, participating memory-maintaining agents update relevant fields via
\begin{equation}
M_c^{(t)}\robusteq
f_{\mathrm{mem}}\left(M_c^{(t-1)},H_c^{(t)},O_c^{(t)},R_c^{(t)}\right).
\label{eq:memory-update}
\end{equation}
where $f_{\mathrm{mem}}$ is implemented through bounded memory-writing Tools that support two field-level operations: \texttt{revise} (correct/replace an existing field) and \texttt{expand} (append newly acquired case information). The lawyer memory functions as a dynamic professional case record (factual main line, evidence ledger, dispute focuses, client communication profile, confirmed positions), while the client memory stores party-side narrative, perceived procedural progress, litigation goals, and bottom line. This separation lets \textsc{\textbf{LegalWorld}} model the gap between professional legal cognition and ordinary party cognition while keeping both consistent across the litigation life cycle.

\subsection{Procedural Skill and Tool Support}
\label{sec:tool-skill-supply}

The Skill/Tool layer provides stage-specific procedural support for agents \citep{schick2023toolformer,qin2024toolllm,wang2024codeact}. In the stage interface, $\Sigma_a^{(t)}$ bundles visible \textbf{Skills} $\mathcal{K}_a^{(t)}$ with executable \textbf{Tools} $\mathcal{T}_a^{(t)}$: Skills specify steps, constraints, and outputs, while Tools handle memory, retrieval, artifacts, export, and citation checks. Stage gating with $V_{a,c}^{(t)}$ and $\mathcal{U}_a^{(t)}$ prevents hidden, ground-truth, or post-stage leakage; details appear in Appendices~\ref{app:tool-skill-catalogue}--\ref{app:skill-fields}.

\begin{table*}[!t]
\centering
\small
\setlength{\tabcolsep}{2.6pt}
\renewcommand{\arraystretch}{0.92}
\begin{tabular*}{\textwidth}{@{\extracolsep{\fill}}lrrrrrr@{}}
\toprule
Target & \multicolumn{1}{c}{\makecell{Procedural /\\Stance}} & \multicolumn{1}{c}{\makecell{Coherence /\\Distinct.}} & \multicolumn{1}{c}{\makecell{Human\\Avg.}} & \multicolumn{1}{c}{\makecell{LLM\\Avg.}} & \multicolumn{1}{c}{\makecell{Mean Diff.\\(H--L)}} & \multicolumn{1}{c}{\makecell{Within\\$\pm$1 (\%)}} \\
\midrule
\rowcolor{tablegroupgray}
\multicolumn{7}{c}{\textbf{Stage Authenticity}} \\
LC & 8.91 & 9.00 & 8.95 & 7.84 & +1.11 & 52.60 \\
CD/DD & 8.85 & 8.95 & 8.90 & \textbf{8.20} & \textbf{+0.70} & \textbf{63.92} \\
FIT & 8.94 & 8.99 & \underline{8.96} & 7.88 & +1.09 & 56.19 \\
AD/AR & 8.92 & 8.99 & \underline{8.96} & \underline{8.18} & \underline{+0.78} & \underline{63.02} \\
SIT & 8.99 & 9.04 & \textbf{9.01} & 7.90 & +1.12 & 48.44 \\
\underline{Overall} & 8.92 & 8.99 & 8.96 & 8.00 & +0.96 & 56.85 \\
\midrule
\rowcolor{tablegroupgray}
\multicolumn{7}{c}{\textbf{Role Consistency}} \\
Client & \textbf{9.09} & 8.84 & \underline{8.96} & 7.73 & +1.23 & 56.70 \\
Lawyer & \underline{9.07} & \textbf{9.01} & \textbf{9.04} & \underline{9.19} & \textbf{-0.15} & \textbf{92.78} \\
Judge & 8.91 & \underline{8.96} & 8.93 & \textbf{9.48} & \underline{-0.55} & \underline{81.44} \\
\underline{Overall} & 9.02 & 8.93 & 8.98 & 8.80 & +0.18 & 76.98 \\
\bottomrule
\end{tabular*}
\caption{Human--LLM agreement validation for \textsc{\textbf{LegalWorld}}. The first two numeric columns are human-average rubric sub-dimensions, not separate annotators: procedural compliance/process coherence for Stage Authenticity, and stance authenticity/role distinguishability for Role Consistency. Mean Diff. is Human minus LLM; Within $\pm$1 is the share of aligned metric pairs within one point. Bold and numeric underlining mark best/second-best non-overall results within each group; underlined Overall rows report group aggregates.}
\label{tab:human-llm-validation}
\vspace{-0.5\baselineskip}
\end{table*}

\section{Experiments}

\subsection{Experimental Setup}

All main-paper experiments run on the cause-balanced \textbf{Light} split defined in Section~\ref{sec:data-foundation}, which keeps the per-case simulation cost tractable while preserving cause coverage. LLM-as-Judge evaluations use Claude-Sonnet-4.6~\citep{anthropic2026claude-sonnet-46-system-card}, while non-evaluated lawyer agents and other environment agents use Qwen3.5-Plus.

Experiments cover two main components: environment reliability---stage authenticity and role consistency (\S\ref{sec:env-validation}) together with judicial output alignment (\S\ref{sec:judicial-alignment}), with cross-stage causal dependence reported in Appendix~\ref{app:cross-stage-dependence}; and cross-model lawyer-backbone benchmarking across the litigation life cycle (\S\ref{sec:cross-model}). We then add a final exploratory probe showing that long-horizon interaction traces produced by a life-cycle environment can serve as training signals for improving legal-agent capabilities (\S\ref{sec:trajectory-reflection}).

\subsection{Environment Reliability Validation}
\label{sec:env-validation}

We validate \textsc{\textbf{LegalWorld}} as a reliable foundation for downstream agent evaluation along two main dimensions---stage authenticity and role consistency---complemented by judicial output alignment (\S\ref{sec:judicial-alignment}) and cross-stage causal dependence (Appendix~\ref{app:cross-stage-dependence}). For both main dimensions, we further compare LLM-as-Judge results with 18,992 individual ratings from 217 legal-background human evaluators.

\paragraph{Stage Authenticity.}
Stage authenticity tests whether simulated trajectories follow legal procedure. Each stage dialogue is scored on a 10-point scale across procedural compliance and process coherence, covering Civil Procedure Law alignment, procedural-step integrity, information transfer, turn-taking, role boundaries, and professional expression. The evaluation covers all stages to obtain stable average score estimates for each stage.

\begin{figure}[!t]
  \centering
  \includegraphics[width=\columnwidth]{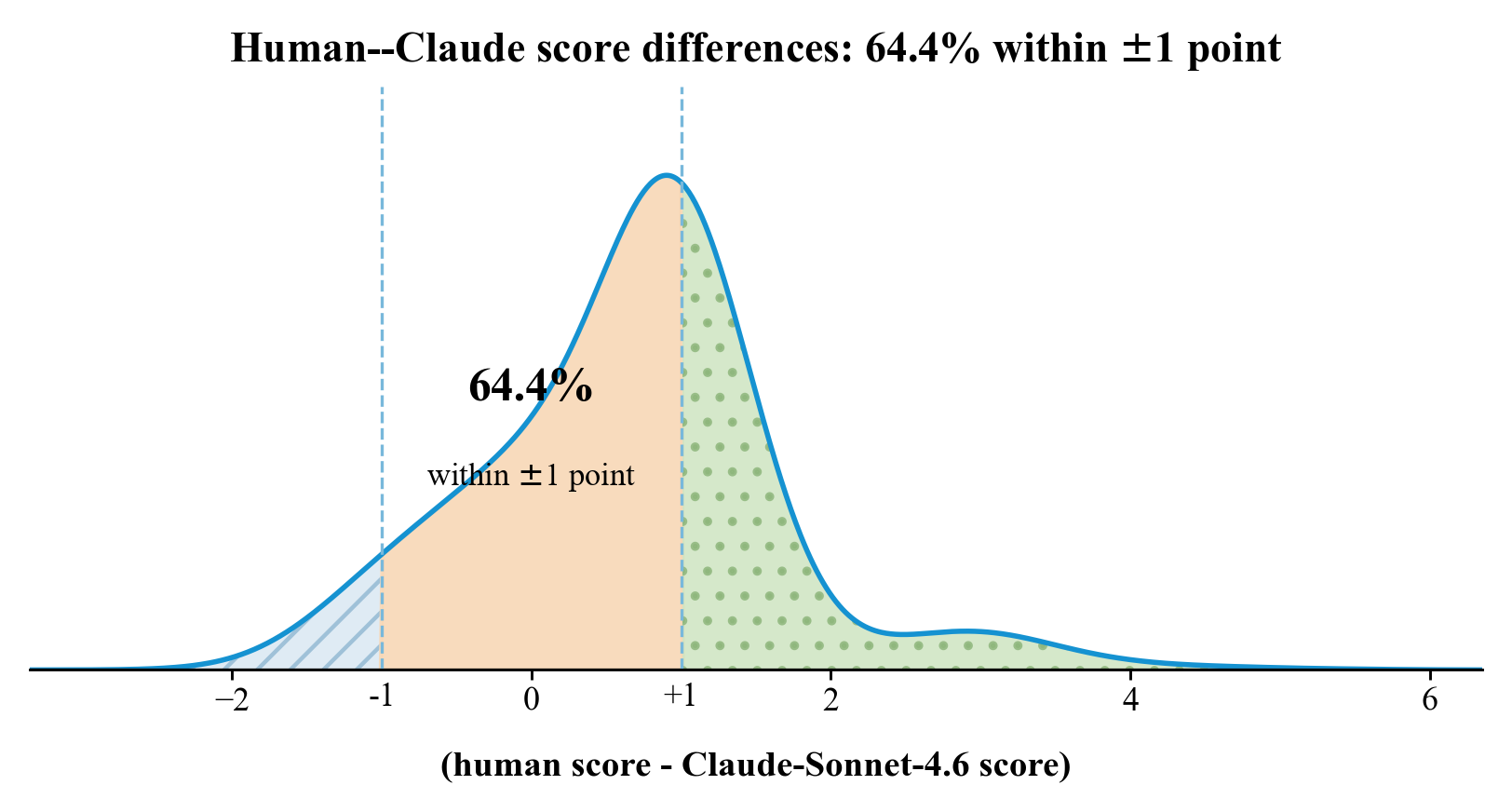}
    \caption{Human minus Claude-Sonnet-4.6 LLM-as-Judge score differences across aligned metric-level pairs. Positive values indicate higher human scores; mean difference is $+0.67$, $\sigma=0.98$, and 64.4\% fall within one point ($|\Delta|\leq 1.0$).}
  \label{fig:human-llm-difference}
\end{figure}

\paragraph{Role Consistency.}
Role consistency checks whether agents maintain coherent behavior across the litigation life cycle. Role behavior is scored on authenticity of stance and motivation, which checks whether behavior conforms to each role's interest position, and inter-role distinguishability, which checks whether clients, lawyers, and judges remain clearly separable.

Table~\ref{tab:human-llm-validation} and Figure~\ref{fig:human-llm-difference} provide the main human-validation evidence. Across all five stages and three roles, 217 legal-background evaluators rate \textsc{\textbf{LegalWorld}} at 8.96/10 on stage authenticity and 8.98/10 on role consistency, indicating that the trajectories are perceived as procedurally faithful and role-coherent. Claude-Sonnet-4.6 applies the same rubric more conservatively ($+0.96$ lower mean on stage authenticity), but still scores all stages in the 7.7--9.5 range, suggesting that the gap mainly reflects rater strictness rather than disagreement about trajectory validity. Role consistency shows tighter agreement (within $\pm 1$ in 77\% of pairs), with the main residual mismatch on the client role, where humans tolerate more legally informed client speech. We therefore use LLM-as-Judge as the primary scorer in \S\ref{sec:cross-model}, treating human ratings as evidence that its scores are conservative lower bounds on environment quality.

\begin{figure}[!t]
  \centering
  \includegraphics[width=\columnwidth]{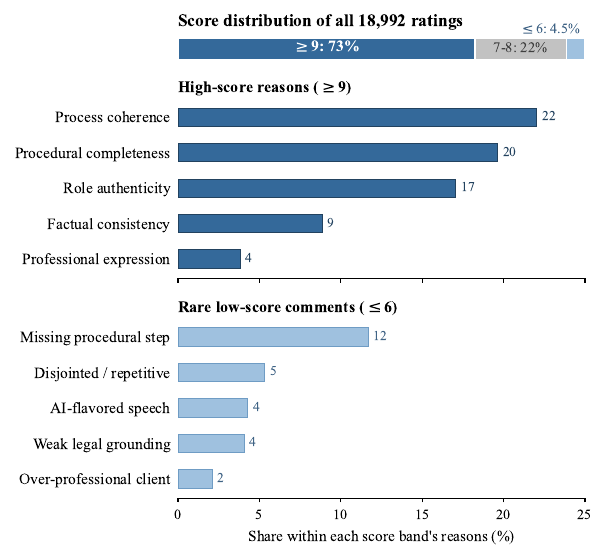}
  \caption{Human rating reason analysis from the 18{,}992 free-text justifications. \emph{Top}: the overall score distribution---73\% of ratings are $\geq 9$ and only 4.5\% are $\leq 6$. \emph{Bottom}: selected informative reason themes summarized separately for the high-score band ($\geq 9$) and the rare low-score band ($\leq 6$); each bar reports a theme's share within its own score band after assigning each justification one quality theme and omitting the uninformative \emph{other} class. High-score reasons mainly reflect process coherence, procedural completeness, and role authenticity, while rare low-score comments indicate localized refinement points.}
  \label{fig:reason-themes}
\end{figure}

\paragraph{Evaluation Reason Analysis.}
Human ratings are overwhelmingly high: 73\% of all 18{,}992 ratings are $\geq 9$ and only 4.5\% are $\leq 6$ (Figure~\ref{fig:reason-themes}, top). We analyze the free-text justifications as a reason-composition check rather than a contrastive error analysis. Among the informative coded themes in the high-score band, the most frequent reasons point to process coherence, procedural completeness, and role authenticity, indicating that evaluators recognized concrete procedural quality in the trajectories. The rare low-score band is summarized separately to identify localized refinement points, with some comments mentioning missing procedural links, repetitive turns, AI-flavored phrasing, or weak legal grounding in particular moments.

\subsection{Judicial Output Alignment}
\label{sec:judicial-alignment}

Beyond process authenticity, we check whether the judgments produced inside \textsc{\textbf{LegalWorld}} match real judicial outputs. A rule-based metric compares each generated first- and second-instance judgment against its real counterpart on six structured elements---verdict, reasoning, legal reference, appeal action, entity, and structure---using set-overlap F1; Appendix~\ref{app:rule-alignment} gives the dimension definitions and scoring formula. Table~\ref{tab:judicial-output-alignment} reports the alignment on a 0--10 scale. Generated judgments align closely with real ones on structure, entity, and factual reasoning, while the largest residual gap is on legal-reference precision, where models tend to cite the correct provision family but not always the exact article. This level of alignment indicates that the environment's judicial outputs are faithful enough to serve as references for downstream evaluation, further supporting the accuracy of \textsc{\textbf{LegalWorld}} as a civil-litigation simulation environment.

\begin{table}[t]
\centering
\footnotesize
\setlength{\tabcolsep}{4pt}
\renewcommand{\arraystretch}{1.12}
\begin{tabularx}{\columnwidth}{@{}Y *{3}{>{\centering\arraybackslash}p{0.16\columnwidth}}@{}}
\toprule
\rowcolor{tablegroupgray}
\multicolumn{4}{c}{\textbf{Judicial Output Alignment}} \\
\midrule
Judgment element & FIT & SIT & Overall \\
\midrule
Verdict & 8.17 & 7.78 & 7.98 \\
Reasoning & 8.22 & 8.69 & 8.45 \\
Legal reference & 6.76 & 7.29 & 7.02 \\
Appeal action & {\NA} & 7.58 & 7.58 \\
Entity & 8.99 & 8.78 & 8.89 \\
Structure & 9.70 & 9.02 & 9.36 \\
\midrule
\underline{Overall} & 8.37 & 8.19 & 8.28 \\
\bottomrule
\end{tabularx}
\caption{Rule-based output-alignment validation for generated judicial judgments against their real counterparts, scored 0--10 over six structured judgment elements. Columns are first-instance (FIT), second-instance (SIT), and their combination (Overall); the underlined bottom row averages across elements. Dimension definitions and the scoring formula are in Appendix~\ref{app:rule-alignment}.}
\label{tab:judicial-output-alignment}
\end{table}

\begin{table*}[!t]
\centering
\footnotesize
\setlength{\tabcolsep}{5pt}
\renewcommand{\arraystretch}{1.2}
\resizebox{\textwidth}{!}{%
\begin{tabular}{@{}lcccccccc@{}}
\toprule
 & Legal Consultation & \multicolumn{4}{c}{Document Drafting} & \multicolumn{3}{c}{Courtroom Advocacy} \\
\cmidrule(lr){2-2}\cmidrule(lr){3-6}\cmidrule(lr){7-9}
Model & \makecell{Issue\\Spotting} & \makecell{Party\\Identification} & \makecell{Claim\\Construction} & \makecell{Fact\\Marshalling} & \makecell{Evidence\\Marshalling} & \makecell{Position\\Consistency} & \makecell{Evidentiary\\Advocacy} & \makecell{Legal\\Reasoning} \\
\midrule
Kimi-K2.5 & \textbf{0.67} & \textbf{0.80}\,/\,\textbf{0.85} & \underline{0.69}\,/\,\underline{0.71} & \textbf{0.68}\,/\,\textbf{0.74} & \textbf{0.74}\,/\,0.52 & \textbf{0.63}\,/\,\textbf{0.65} & \underline{0.55}\,/\,\textbf{0.59} & \underline{0.56}\,/\,\textbf{0.58} \\
Qwen3.5-Plus & 0.62 & \underline{0.71}\,/\,\textbf{0.85} & \textbf{0.72}\,/\,\textbf{0.72} & 0.66\,/\,0.70 & 0.69\,/\,\underline{0.70} & \underline{0.62}\,/\,\underline{0.64} & 0.53\,/\,0.54 & 0.53\,/\,\textbf{0.58} \\
GPT-5.2 & \underline{0.63} & 0.53\,/\,\underline{0.83} & 0.60\,/\,0.70 & 0.60\,/\,0.68 & 0.71\,/\,0.63 & \underline{0.62}\,/\,\underline{0.64} & \textbf{0.61}\,/\,\underline{0.55} & \textbf{0.57}\,/\,\underline{0.57} \\
DeepSeek-V4-Flash & 0.62 & 0.64\,/\,\underline{0.83} & 0.65\,/\,0.65 & \underline{0.67}\,/\,0.69 & 0.69\,/\,0.57 & 0.60\,/\,0.63 & 0.52\,/\,0.54 & 0.52\,/\,\underline{0.57} \\
GLM-4.7 & 0.54 & 0.56\,/\,0.82 & 0.66\,/\,0.66 & \underline{0.67}\,/\,0.70 & \underline{0.72}\,/\,\textbf{0.71} & 0.61\,/\,0.61 & 0.53\,/\,0.52 & 0.48\,/\,0.51 \\
Qwen3.5-Flash & 0.56 & 0.46\,/\,0.82 & 0.56\,/\,0.69 & 0.62\,/\,\underline{0.72} & 0.50\,/\,0.53 & 0.53\,/\,0.55 & 0.46\,/\,0.41 & 0.45\,/\,0.45 \\
\bottomrule
\end{tabular}}
\caption{Cross-model task-capability profile on LongJud-Bench. Rows are lawyer backbones; columns are eight legal capabilities grouped by litigation phase. Except for the consultation capability, each cell reports first-instance\,/\,second-instance scores in $[0,1]$: for \emph{document drafting} these come from the first-instance (CD/DD) and second-instance (AD/AR) pleadings, and for \emph{courtroom advocacy} from the first- and second-instance trials. \textbf{Bold} and \underline{underline} mark the best and second-best backbone on each side independently.}
\label{tab:cross-model}
\end{table*}

\begin{figure}[!t]
  \centering
  \includegraphics[width=\columnwidth]{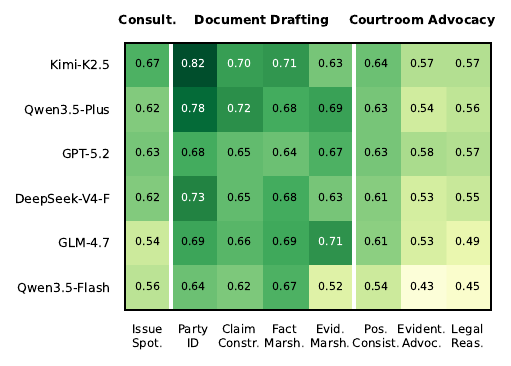}
  \caption{Capability heatmap of the six backbones (consultation uses its single score; paired drafting and advocacy cells use the mean of the first-/second-instance scores in Table~\ref{tab:cross-model}; darker is higher). The courtroom-advocacy rows stay lighter than the drafting rows across all backbones, marking advocacy as the shared frontier.}
  \label{fig:cross-model-capability-heatmap}
\end{figure}

\subsection{LongJud-Bench Evaluation Framework}

Building on the validated environment, LongJud-Bench scores the target lawyer agent $a^\ast$ over the complete litigation process through eight legal capabilities grouped by litigation phase: \emph{legal consultation} (issue spotting); \emph{document drafting} (party identification, claim construction, fact marshalling, and evidence marshalling); and \emph{courtroom advocacy} (position consistency, evidentiary advocacy, and legal reasoning). This capability-level view aligns evaluation with the professional functions a litigation lawyer must perform across the full life cycle.

Each capability is evaluated with either rule-based matching or LLM-as-Judge scoring \citep{zheng2023judging}, depending on the evidence type. Consultation is scored question-by-question against reference answers grounded in case facts and applicable statutes. Drafting capabilities combine exact match for structured party slots with 0--10 semantic scoring for claims or defenses, facts, and evidence in the first- and second-instance pleadings. Courtroom-advocacy capabilities apply multi-dimensional 0--10 scoring to $a^\ast$'s trial statements, covering consistency with the pleaded position, fact-and-evidence use, and legal reasoning. Every item is normalized to $[0,1]$; the per-capability formulas and the full scoring-item-to-capability mapping are given in Appendix~\ref{app:longjud-bench-details}.

\subsection{Cross-Model Capability Profile}
\label{sec:cross-model}

We instantiate each backbone as the target lawyer agent in \textsc{\textbf{LegalWorld}} while fixing the surrounding roles to Qwen3.5-Plus, and read out the eight capabilities of Table~\ref{tab:cross-model} across the first- and second-instance sides of the life cycle. Three patterns stand out.

\paragraph{No backbone wins everywhere.} Models that look comparable in aggregate diverge sharply once the trajectory is decomposed by capability. Kimi-K2.5 is strongest on the drafting capabilities and on keeping courtroom advocacy aligned with the pleaded position, whereas GPT-5.2~\citep{openai2025gpt52-system-card}---weaker at the formal drafting slots---leads precisely where it matters most in court, on first-instance evidentiary advocacy and legal reasoning; Qwen3.5-Plus is in turn the strongest claim constructor. These trade-offs are invisible to any single aggregate score and are exactly what a capability profile is meant to surface.

\paragraph{Courtroom advocacy is the frontier.} Across all backbones the three advocacy capabilities---position consistency, evidentiary advocacy, and legal reasoning---sit well below the drafting capabilities (Figure~\ref{fig:cross-model-capability-heatmap}), and the gap widens for the weaker models. Multi-turn courtroom advocacy, where the lawyer must integrate accumulated memory, opposing statements, and judge prompts on the fly, remains the hardest competency for current models and the most discriminative target for future legal-agent training.

\paragraph{Formal sub-skills saturate while reasoning discriminates.} Structural competencies such as party identification are near-saturated on the second-instance side, where the first-instance judgment scaffolds the document, so they barely separate backbones; the discriminative signal concentrates in evidentiary advocacy and legal reasoning. The first-to-second-instance shift is itself informative---most capabilities improve once the first-instance judgment is available as scaffolding, but evidence marshalling can instead fall on appeal, where marshalling new evidence is harder than reusing an established record.

\subsection{Trajectory Reflection as an Exploratory Training Signal}
\label{sec:trajectory-reflection}

Beyond benchmarking, \textsc{\textbf{LegalWorld}} produces complete procedural traces---dialogues, drafted artifacts, judgments, memory updates, and evaluation signals---that can be reused as grounded experience for training later agents. We do not treat reflection as part of the core environment framework; instead, we run a lightweight probe called \emph{Reflective Legal Skill} (RLS) to test whether the generated long-horizon data contains reusable legal-practice signal.

RLS is produced in two steps. First, after a case finishes, we build a post-case reflection context from the visible case materials, lawyer actions, generated artifacts, memory updates, and evaluation signals, and ask the lawyer agent to summarize the completed trajectory into a candidate reusable rule. Second, the candidate is checked against existing cause-matched Skills for overlap and redundancy, and is retained only if it specifies a reusable trigger condition, role or stage scope, procedural correction principle, and expected-output constraint. The retained rule becomes an optional cause-matched Skill note. In later same-cause cases, the same baseline lawyer agent receives the note as an additional Skill; the case seed, Tools, and in-case memory mechanism remain unchanged.

Table~\ref{tab:reflective-skill-growth} reports the result on the three most frequent civil causes of action in our dataset: post-divorce property disputes, private lending, and labor disputes. Adding these simple reflective Skill notes raises the average LongJud-Bench overall score from 61.56 to 65.29 (+3.73 points). Gains appear on all three causes, with larger improvements on reflected cases (+4.20 on average) and still positive transfer to held-out same-cause cases not used to write the Skill (+2.34 on average). This exploratory result suggests that life-cycle interaction traces are useful not only for evaluation, but also as procedurally grounded data for improving legal agents.

\begin{table}[H]
\centering
\footnotesize
\setlength{\tabcolsep}{1.2pt}
\renewcommand{\arraystretch}{1.04}
\begin{tabular*}{\columnwidth}{@{\extracolsep{\fill}}>{\raggedright\arraybackslash}p{0.28\columnwidth}>{\centering\arraybackslash}p{0.115\columnwidth}>{\centering\arraybackslash}p{0.10\columnwidth}>{\centering\arraybackslash}p{0.12\columnwidth}>{\centering\arraybackslash}p{0.11\columnwidth}>{\centering\arraybackslash}p{0.13\columnwidth}@{}}
\toprule
& \multicolumn{2}{c}{\textbf{Score}} & \multicolumn{3}{c}{\textbf{Diff}} \\
\cmidrule(lr){2-3}\cmidrule(lr){4-6}
Cause & Base. & RLS & Overall & Refl. & Held-out \\
\midrule
\makecell[l]{Post-divorce\\property} & \underline{60.08} & \underline{63.76} & \underline{+3.69} & \underline{+3.81} & \textbf{+3.29} \\
Private lending & 58.56 & 61.84 & +3.28 & +3.63 & \underline{+2.32} \\
Labor dispute & \textbf{66.25} & \textbf{70.47} & \textbf{+4.22} & \textbf{+5.24} & +1.42 \\
\midrule
\underline{Average} & 61.56 & 65.29 & +3.73 & +4.20 & +2.34 \\
\bottomrule
\end{tabular*}
\caption{Exploratory RLS gains across high-frequency civil causes. Scores are LongJud-Bench overall scores on a 0--100 scale; Base. is the same lawyer agent without reflective Skills. Refl. measures cases used to produce the reflective note, while Held-out measures same-cause cases not used for reflection.}
\label{tab:reflective-skill-growth}
\end{table}

\section{Related Work}

\paragraph{Legal simulation and generative agents.} LLM-based social simulation elicits coherent role behavior and long-horizon interaction \citep{park2023generative,wang2024survey,li_simulating_2025}, with extensions to professional workflows \citep{li_agent_2025,jin_evolution_2025} and persona-driven diversification \citep{tseng-etal-2024-two}. Existing legal simulators remain narrower than the full litigation process: AgentCourt and AgentsCourt model adversarial trial procedures \citep{chen_agentcourt_nodate,he-etal-2024-agentscourt-emnlp}, Ready Jurist One covers multiple scenarios but initializes each from shared case ground truth \citep{jia2026readyjuristonebenchmarking}, and Law in Silico studies socio-legal dynamics through group simulation \citep{wang_law_2025}. \textsc{\textbf{LegalWorld}} differs by chaining consultation, drafting, and both trial instances into a single life cycle, so that factual carryover and error amplification become observable within one case.

\paragraph{Legal capability benchmarks.} Existing legal AI benchmarks mostly measure local capabilities such as statute retrieval, document generation, single-case reasoning, and outcome prediction \citep{fei-etal-2024-lawbench,guha2023legalbench,xiao2018cail,zhong2018overview,li-etal-2024-lexeval,deng-etal-2024-learning,gao-etal-2024-enhancing-legal}. Long-context benchmarks test single-pass input handling \citep{bai-etal-2024-longbench,bai-etal-2025-longbench}, while agent-memory work focuses on persistence across long conversations \citep{maharana-etal-2024-evaluating}. LongJud-Bench instead evaluates consultation, drafting, trial advocacy, appeal, and second-instance trial as connected stages of one case, measuring local quality together with cross-stage error propagation.




\section{Conclusion}

We presented \textsc{\textbf{LegalWorld}}, a life-cycle interactive environment for Chinese civil litigation grounded in 75,309 paired civil judgments and equipped with reusable infrastructure for long-horizon agents, which turns each dispute into a connected trajectory across consultation, drafting, and two trial instances. Building on this foundation, we constructed \textbf{LongJud-Bench} to evaluate legal-agent capability across the full procedural life cycle.

Two implications follow. First, trajectory-level evaluation exposes cross-stage causal dependence that single-stage benchmarks cannot detect, framing legal-agent capability as a trajectory-level property rather than a collection of isolated subtask scores. Second, beyond evaluation, the life-cycle interaction traces produced by \textsc{\textbf{LegalWorld}}---legal artifacts, multi-role dialogues, and cross-stage memory updates---are themselves procedurally grounded data for agent improvement: our lightweight trajectory-reflection probe shows that even simple post-case reflection can improve later same-cause legal-agent behavior.

\FloatBarrier
\clearpage

\section*{Limitations}

This work focuses on Chinese civil litigation and paired first-/second-instance judgment data, so the current environment does not yet cover criminal, administrative, enforcement, or retrial procedures. The simulation also simplifies exceptional procedural events and relies on benchmark scoring rather than real legal service outcomes. Future work should extend the life-cycle formulation to other procedures, incorporate branching events such as jurisdictional objections, preservation applications, counterclaims, expert opinions, and settlement failures, and validate human-agent collaboration with legal professionals.

\section*{Ethics Statement}

All judgment data used in this work come from public legal sources and are processed for research and evaluation. Because public judgments may still contain party names, case numbers, addresses, organization names, or other legally relevant identifiers, we remove or anonymize direct personal identifiers when they are not required for benchmark construction or reproducible evaluation. \textsc{\textbf{LegalWorld}} and LongJud-Bench are intended for legal AI simulation, benchmarking, and training support, not for replacing lawyers or judges or making real legal decisions. Model outputs may contain legal errors or unsupported reasoning, so any deployment-facing use should include professional review, privacy protection, and clear disclosure that the system is an AI research tool.

The human-rating study used public legal-case materials and did not collect personally identifiable information from evaluators. The 217 legal-background evaluators were informed of the research purpose, participated knowingly and voluntarily, and were told that their 18,992 ratings would be analyzed only in aggregate. The study did not intervene in real legal disputes or collect private party data beyond information already available from public legal sources.

\nolinenumbers
\bibliography{arxiv_version_reference}

\begin{thebibliography}{35}
\providecommand{\natexlab}[1]{#1}

\bibitem[{{Anthropic}(2026)}]{anthropic2026claude-sonnet-46-system-card}
{Anthropic}. 2026.
\newblock \href {https://anthropic.com/claude-sonnet-4-6-system-card} {Claude
  sonnet 4.6 system card}.
\newblock System card.

\bibitem[{Bai et~al.(2024)Bai, Lv, Zhang, Lyu, Tang, Huang, Du, Liu, Zeng, Hou,
  Dong, Tang, and Li}]{bai-etal-2024-longbench}
Yushi Bai, Xin Lv, Jiajie Zhang, Hongchang Lyu, Jiankai Tang, Zhidian Huang,
  Zhengxiao Du, Xiao Liu, Aohan Zeng, Lei Hou, Yuxiao Dong, Jie Tang, and
  Juanzi Li. 2024.
\newblock \href {https://doi.org/10.18653/v1/2024.acl-long.172} {{LongBench}: A
  bilingual, multitask benchmark for long context understanding}.
\newblock In \emph{Proceedings of the 62nd Annual Meeting of the Association
  for Computational Linguistics (Volume 1: Long Papers)}, pages 3119--3137.
  Association for Computational Linguistics.

\bibitem[{Bai et~al.(2025)Bai, Tu, Zhang, Peng, Wang, Lv, Cao, Xu, Hou, Dong,
  Tang, and Li}]{bai-etal-2025-longbench}
Yushi Bai, Shangqing Tu, Jiajie Zhang, Hao Peng, Xiaozhi Wang, Xin Lv, Shulin
  Cao, Jiazheng Xu, Lei Hou, Yuxiao Dong, Jie Tang, and Juanzi Li. 2025.
\newblock \href {https://doi.org/10.18653/v1/2025.acl-long.183} {{L}ong{B}ench
  v2: Towards deeper understanding and reasoning on realistic long-context
  multitasks}.
\newblock In \emph{Proceedings of the 63rd Annual Meeting of the Association
  for Computational Linguistics (Volume 1: Long Papers)}, pages 3639--3664,
  Vienna, Austria. Association for Computational Linguistics.

\bibitem[{Chen et~al.(2025)Chen, Fan, Gong, Xie, Li, Liu, Li, Qu,
  Alinejad-Rokny, Ni, and Yang}]{chen_agentcourt_nodate}
Guhong Chen, Liyang Fan, Zihan Gong, Nan Xie, Zixuan Li, Ziqiang Liu, Chengming
  Li, Qiang Qu, Hamid Alinejad-Rokny, Shiwen Ni, and Min Yang. 2025.
\newblock \href {https://doi.org/10.18653/v1/2025.findings-acl.304}
  {{AgentCourt}: Simulating court with adversarial evolvable lawyer agents}.
\newblock In \emph{Findings of the Association for Computational Linguistics:
  ACL 2025}, pages 5850--5865, Vienna, Austria. Association for Computational
  Linguistics.

\bibitem[{Cui et~al.(2024)Cui, Ning, Li, Chen, Yan, Li, Ling, Tian, and
  Yuan}]{cui2024chatlaw}
Jiaxi Cui, Munan Ning, Zongjian Li, Bohua Chen, Yang Yan, Hao Li, Bin Ling,
  Yonghong Tian, and Li~Yuan. 2024.
\newblock \href {https://arxiv.org/abs/2306.16092} {Chatlaw: A multi-agent
  collaborative legal assistant with knowledge graph enhanced
  mixture-of-experts large language model}.
\newblock \emph{Preprint}, arXiv:2306.16092.

\bibitem[{{DeepSeek-AI}(2025)}]{deepseek2025v32}
{DeepSeek-AI}. 2025.
\newblock \href {https://arxiv.org/abs/2512.02556} {Deepseek-v3.2: Pushing the
  frontier of open large language models}.
\newblock \emph{arXiv preprint arXiv:2512.02556}.

\bibitem[{Deng et~al.(2024)Deng, Mao, and Dou}]{deng-etal-2024-learning}
Chenlong Deng, Kelong Mao, and Zhicheng Dou. 2024.
\newblock \href {https://doi.org/10.18653/v1/2024.emnlp-main.73} {Learning
  interpretable legal case retrieval via knowledge-guided case reformulation}.
\newblock In \emph{Proceedings of the 2024 Conference on Empirical Methods in
  Natural Language Processing}, pages 1253--1265, Miami, Florida, USA.
  Association for Computational Linguistics.

\bibitem[{Fei et~al.(2024)Fei, Shen, Zhu, Zhou, Han, Huang, Zhang, Chen, Yin,
  Shen, Ge, and Ng}]{fei-etal-2024-lawbench}
Zhiwei Fei, Xiaoyu Shen, Dawei Zhu, Fengzhe Zhou, Zhuo Han, Alan Huang,
  Songyang Zhang, Kai Chen, Zhixin Yin, Zongwen Shen, Jidong Ge, and Vincent
  Ng. 2024.
\newblock \href {https://doi.org/10.18653/v1/2024.emnlp-main.452}
  {{L}aw{B}ench: Benchmarking legal knowledge of large language models}.
\newblock In \emph{Proceedings of the 2024 Conference on Empirical Methods in
  Natural Language Processing}, pages 7933--7962, Miami, Florida, USA.
  Association for Computational Linguistics.

\bibitem[{Gao et~al.(2024)Gao, Xiao, Liu, Chen, Liu, and
  Sun}]{gao-etal-2024-enhancing-legal}
Cheng Gao, Chaojun Xiao, Zhenghao Liu, Huimin Chen, Zhiyuan Liu, and Maosong
  Sun. 2024.
\newblock \href {https://doi.org/10.18653/v1/2024.emnlp-main.402} {Enhancing
  legal case retrieval via scaling high-quality synthetic query-candidate
  pairs}.
\newblock In \emph{Proceedings of the 2024 Conference on Empirical Methods in
  Natural Language Processing}, pages 7086--7100, Miami, Florida, USA.
  Association for Computational Linguistics.

\bibitem[{Guha et~al.(2023)Guha, Nyarko, Ho, R{\'e}, Chilton, Narayana,
  Chohlas-Wood, Peters, Waldon, Rockmore, Zambrano, Talisman, Hoque, Surani,
  Fagan, Sarfaty, Dickinson, Porat, Hegland, Wu, Nudell, Niklaus, Nay, Choi,
  Tobia, Hagan, Ma, Livermore, Rasumov-Rahe, Holzenberger, Kolt, Henderson,
  Rehaag, Goel, Gao, Williams, Gandhi, Zur, Iyer, and Li}]{guha2023legalbench}
Neel Guha, Julian Nyarko, Daniel~E. Ho, Christopher R{\'e}, Adam Chilton,
  Aditya Narayana, Alex Chohlas-Wood, Austin Peters, Brandon Waldon, Daniel~N.
  Rockmore, Diego Zambrano, Dmitry Talisman, Enam Hoque, Faiz Surani, Frank
  Fagan, Galit Sarfaty, Gregory~M. Dickinson, Haggai Porat, Jason Hegland, and
  21 others. 2023.
\newblock \href {https://openreview.net/forum?id=WqSPQFxFRC} {{LegalBench}: A
  collaboratively built benchmark for measuring legal reasoning in large
  language models}.
\newblock In \emph{Advances in Neural Information Processing Systems 36
  (NeurIPS 2023) Datasets and Benchmarks Track}.

\bibitem[{He et~al.(2024)He, Cao, Wang, Jin, Chen, Xu, Li, Liu, and
  Zhao}]{he-etal-2024-agentscourt-emnlp}
Zhitao He, Pengfei Cao, Chenhao Wang, Zhuoran Jin, Yubo Chen, Jiexin Xu,
  Huaijun Li, Kang Liu, and Jun Zhao. 2024.
\newblock \href {https://doi.org/10.18653/v1/2024.findings-emnlp.549}
  {{AgentsCourt}: Building judicial decision-making agents with court debate
  simulation and legal knowledge augmentation}.
\newblock In \emph{Findings of the Association for Computational Linguistics:
  EMNLP 2024}, pages 9399--9416, Miami, Florida, USA. Association for
  Computational Linguistics.

\bibitem[{Jia et~al.(2026)Jia, Yue, Chen, Wang, Liu, Li, Song, and
  Wei}]{jia2026readyjuristonebenchmarking}
Zheng Jia, Shengbin Yue, Wei Chen, Siyuan Wang, Yidong Liu, Zejun Li, Yun Song,
  and Zhongyu Wei. 2026.
\newblock \href {https://arxiv.org/abs/2507.04037} {Ready jurist one:
  Benchmarking language agents for legal intelligence in dynamic environments}.
\newblock \emph{Preprint}, arXiv:2507.04037.

\bibitem[{Jin et~al.(2025)Jin, Wang, Gao, Yang, Chunjia, and
  Wang}]{jin_evolution_2025}
Sheng Jin, Haoming Wang, Zhiqi Gao, Yongbo Yang, Bao Chunjia, and Chengliang
  Wang. 2025.
\newblock \href {https://doi.org/10.48550/arXiv.2510.11290} {Evolution in
  simulation: {AI}-agent school with dual memory for high-fidelity educational
  dynamics}.
\newblock \emph{Preprint}, arxiv:2510.11290 [cs].

\bibitem[{Kyung et~al.(2025)Kyung, Chung, Bae, Kim, Sohn, Kim, Kim, and
  Choi}]{patientsim2025physionet}
Daeun Kyung, Hyunseung Chung, Seongsu Bae, Jiho Kim, Jae~Ho Sohn, Taerim Kim,
  Soo~Kyung Kim, and Edward Choi. 2025.
\newblock \href {https://openreview.net/forum?id=1THAjdP4QJ} {{PatientSim}: A
  persona-driven simulator for realistic doctor-patient interactions}.
\newblock In \emph{Advances in Neural Information Processing Systems 39
  (NeurIPS 2025) Datasets and Benchmarks Track}.

\bibitem[{Li et~al.(2025{\natexlab{a}})Li, Wu, Mo, Qu, Tang, Zhao, Gan, Fan,
  Yu, Zhao, Liang, Alonso, and Larson}]{li_simulating_2025}
Chance~Jiajie Li, Jiayi Wu, Zhenze Mo, Ao~Qu, Yuhan Tang, Kaiya~Ivy Zhao, Yulu
  Gan, Jie Fan, Jiangbo Yu, Jinhua Zhao, Paul Liang, Luis Alonso, and Kent
  Larson. 2025{\natexlab{a}}.
\newblock \href {https://doi.org/10.48550/arXiv.2506.06958} {Simulating society
  requires simulating thought}.
\newblock \emph{Preprint}, arxiv:2506.06958 [cs].

\bibitem[{Li et~al.(2024)Li, Chen, Ai, Wu, Zhang, and
  Liu}]{li-etal-2024-lexeval}
Haitao Li, You Chen, Qingyao Ai, Yueyue Wu, Ruizhe Zhang, and Yiqun Liu. 2024.
\newblock \href {https://doi.org/10.52202/079017-0790} {{LexEval}: A
  comprehensive {C}hinese legal benchmark for evaluating large language
  models}.
\newblock In \emph{Advances in Neural Information Processing Systems 38
  (NeurIPS 2024) Datasets and Benchmarks Track}.

\bibitem[{Li et~al.(2025{\natexlab{b}})Li, Lai, Li, Ren, Zhang, Kang, Wang, Li,
  Zhang, Ma, and Liu}]{li_agent_2025}
Junkai Li, Yunghwei Lai, Weitao Li, Jingyi Ren, Meng Zhang, Xinhui Kang, Siyu
  Wang, Peng Li, Ya-Qin Zhang, Weizhi Ma, and Yang Liu. 2025{\natexlab{b}}.
\newblock \href {https://doi.org/10.48550/arXiv.2405.02957} {Agent hospital: A
  simulacrum of hospital with evolvable medical agents}.
\newblock \emph{Preprint}, arxiv:2405.02957 [cs].

\bibitem[{Liu et~al.(2026)Liu, Zhang, Ma, Deng, Zhu, Li, Li, Shen, and
  Du}]{liu_llm_2026}
Shuang Liu, Ruijia Zhang, Ruoyun Ma, Yujia Deng, Lanyi Zhu, Jiayu Li, Zelong
  Li, Zhibin Shen, and Mengnan Du. 2026.
\newblock \href {https://doi.org/10.48550/arXiv.2601.06216} {{LLM} agents in
  law: Taxonomy, applications, and challenges}.
\newblock \emph{Preprint}, arxiv:2601.06216 [cs].

\bibitem[{Maharana et~al.(2024)Maharana, Lee, Tulyakov, Bansal, Barbieri, and
  Fang}]{maharana-etal-2024-evaluating}
Adyasha Maharana, Dong-Ho Lee, Sergey Tulyakov, Mohit Bansal, Francesco
  Barbieri, and Yuwei Fang. 2024.
\newblock \href {https://aclanthology.org/2024.acl-long.747/} {Evaluating very
  long-term conversational memory of {LLM} agents}.
\newblock In \emph{Proceedings of the 62nd Annual Meeting of the Association
  for Computational Linguistics (Volume 1: Long Papers)}, pages 13851--13870.
  Association for Computational Linguistics.

\bibitem[{{OpenAI}(2025)}]{openai2025gpt52-system-card}
{OpenAI}. 2025.
\newblock \href
  {https://cdn.openai.com/pdf/3a4153c8-c748-4b71-8e31-aecbde944f8d/oai_5_2_system-card.pdf}
  {Update to gpt-5 system card: Gpt-5.2}.
\newblock System card update.

\bibitem[{Packer et~al.(2024)Packer, Wooders, Lin, Fang, Patil, Stoica, and
  Gonzalez}]{packer2024memgpt}
Charles Packer, Sarah Wooders, Kevin Lin, Vivian Fang, Shishir~G. Patil, Ion
  Stoica, and Joseph~E. Gonzalez. 2024.
\newblock \href {https://arxiv.org/abs/2310.08560} {{MemGPT}: Towards {LLMs} as
  operating systems}.
\newblock \emph{Preprint}, arXiv:2310.08560.

\bibitem[{Park et~al.(2023)Park, O'Brien, Cai, Morris, Liang, and
  Bernstein}]{park2023generative}
Joon~Sung Park, Joseph~C. O'Brien, Carrie~J. Cai, Meredith~Ringel Morris, Percy
  Liang, and Michael~S. Bernstein. 2023.
\newblock \href {https://doi.org/10.1145/3586183.3606763} {Generative agents:
  Interactive simulacra of human behavior}.
\newblock In \emph{Proceedings of the 36th Annual ACM Symposium on User
  Interface Software and Technology (UIST '23)}, New York, NY, USA. Association
  for Computing Machinery.

\bibitem[{Qin et~al.(2024)Qin, Liang, Ye, Zhu, Yan, Lu, Lin, Cong, Tang, Qian,
  Zhao, Hong, Tian, Xie, Zhou, Gerstein, Li, Liu, and Sun}]{qin2024toolllm}
Yujia Qin, Shihao Liang, Yining Ye, Kunlun Zhu, Lan Yan, Yaxi Lu, Yankai Lin,
  Xin Cong, Xiangru Tang, Bill Qian, Sihan Zhao, Lauren Hong, Runchu Tian,
  Ruobing Xie, Jie Zhou, Mark Gerstein, Dahai Li, Zhiyuan Liu, and Maosong Sun.
  2024.
\newblock \href {https://openreview.net/forum?id=dHng2O0Jjr} {{ToolLLM}:
  Facilitating large language models to master 16000+ real-world {APIs}}.
\newblock In \emph{The Twelfth International Conference on Learning
  Representations (ICLR)}.

\bibitem[{Ranjan and Ma(2024)}]{ranjan_motivations_nodate}
Riya Ranjan and Megan Ma. 2024.
\newblock \href {https://neurips.cc/virtual/2024/104203} {Motivations for
  reframing large language model benchmarking for legal applications}.
\newblock In \emph{Proceedings of the NeurIPS 2024 Workshop on Evaluating
  Evaluations: Examining Best Practices for Measuring Broader Impacts of
  Generative AI}.

\bibitem[{Schick et~al.(2023)Schick, Dwivedi-Yu, Dess{\`i}, Raileanu, Lomeli,
  Zettlemoyer, Cancedda, and Scialom}]{schick2023toolformer}
Timo Schick, Janne Dwivedi-Yu, Roberto Dess{\`i}, Roberta Raileanu, Maria
  Lomeli, Luke Zettlemoyer, Nicola Cancedda, and Thomas Scialom. 2023.
\newblock \href {https://openreview.net/forum?id=Yacmpz84TH} {Toolformer:
  Language models can teach themselves to use tools}.
\newblock In \emph{Advances in Neural Information Processing Systems 36
  (NeurIPS 2023)}.

\bibitem[{Tseng et~al.(2024)Tseng, Huang, Hsiao, Chen, Huang, Meng, and
  Chen}]{tseng-etal-2024-two}
Yu-Min Tseng, Yu-Chao Huang, Teng-Yun Hsiao, Wei-Lin Chen, Chao-Wei Huang,
  Yu~Meng, and Yun-Nung Chen. 2024.
\newblock \href {https://doi.org/10.18653/v1/2024.findings-emnlp.969} {Two
  tales of persona in {LLMs}: A survey of role-playing and personalization}.
\newblock In \emph{Findings of the Association for Computational Linguistics:
  EMNLP 2024}, pages 16612--16631, Miami, Florida, USA. Association for
  Computational Linguistics.

\bibitem[{Wang et~al.(2024{\natexlab{a}})Wang, Ma, Feng, Zhang, Yang, Zhang,
  Chen, Tang, Chen, Lin, Zhao, Wei, and Wen}]{wang2024survey}
Lei Wang, Chen Ma, Xueyang Feng, Zeyu Zhang, Hao Yang, Jingsen Zhang, Zhiyuan
  Chen, Jiakai Tang, Xu~Chen, Yankai Lin, Wayne~Xin Zhao, Zhewei Wei, and
  Ji-Rong Wen. 2024{\natexlab{a}}.
\newblock \href {https://doi.org/10.1007/s11704-024-40231-1} {A survey on large
  language model based autonomous agents}.
\newblock \emph{Frontiers of Computer Science}, arXiv:2308.11432.

\bibitem[{Wang et~al.(2024{\natexlab{b}})Wang, Chen, Yuan, Zhang, Li, Peng, and
  Ji}]{wang2024codeact}
Xingyao Wang, Yangyi Chen, Lifan Yuan, Yizhe Zhang, Yunzhu Li, Hao Peng, and
  Heng Ji. 2024{\natexlab{b}}.
\newblock \href {https://proceedings.mlr.press/v235/wang24h.html} {Executable
  code actions elicit better {LLM} agents}.
\newblock In \emph{Proceedings of the 41st International Conference on Machine
  Learning (ICML)}, pages 50208--50232. PMLR.

\bibitem[{Wang et~al.(2025)Wang, Chen, Meng, Chen, Yang, and
  Zhang}]{wang_law_2025}
Yiding Wang, Yuxuan Chen, Fanxu Meng, Xifan Chen, Xiaolei Yang, and Muhan
  Zhang. 2025.
\newblock \href {https://doi.org/10.48550/arXiv.2510.24442} {Law in silico:
  Simulating legal society with {LLM}-based agents}.
\newblock \emph{Preprint}, arxiv:2510.24442 [cs].

\bibitem[{Xiao et~al.(2018)Xiao, Zhong, Guo, Tu, Liu, Sun, Feng, Han, Hu, Wang,
  and Xu}]{xiao2018cail}
Chaojun Xiao, Haoxi Zhong, Zhipeng Guo, Cunchao Tu, Zhiyuan Liu, Maosong Sun,
  Yansong Feng, Xianpei Han, Zhen Hu, Heng Wang, and Jianfeng Xu. 2018.
\newblock \href {https://arxiv.org/abs/1807.02478} {{CAIL2018}: A large-scale
  legal dataset for judgment prediction}.
\newblock \emph{Preprint}, arXiv:1807.02478.

\bibitem[{Yue et~al.(2023)Yue, Chen, Wang, Li, Shen, Liu, Zhou, Xiao, Yun,
  Huang, and Wei}]{yue2023disclawllm}
Shengbin Yue, Wei Chen, Siyuan Wang, Bingxuan Li, Chenchen Shen, Shujun Liu,
  Yuxuan Zhou, Yao Xiao, Song Yun, Xuanjing Huang, and Zhongyu Wei. 2023.
\newblock \href {https://arxiv.org/abs/2309.11325} {{DISC}-{LawLLM}:
  Fine-tuning large language models for intelligent legal services}.
\newblock \emph{Preprint}, arXiv:2309.11325.

\bibitem[{Zhang et~al.(2026)Zhang, Long, Bao, Feng, Zhang, Yue, and
  Wang}]{zhang_memskill_2026}
Haozhen Zhang, Quanyu Long, Jianzhu Bao, Tao Feng, Weizhi Zhang, Haodong Yue,
  and Wenya Wang. 2026.
\newblock \href {https://doi.org/10.48550/arXiv.2602.02474} {{MemSkill}:
  Learning and evolving memory skills for self-evolving agents}.
\newblock \emph{Preprint}, arxiv:2602.02474 [cs].

\bibitem[{Zheng et~al.(2023)Zheng, Chiang, Sheng, Zhuang, Wu, Zhuang, Lin, Li,
  Li, Xing, Zhang, Gonzalez, and Stoica}]{zheng2023judging}
Lianmin Zheng, Wei-Lin Chiang, Ying Sheng, Siyuan Zhuang, Zhanghao Wu, Yonghao
  Zhuang, Zi~Lin, Zhuohan Li, Dacheng Li, Eric~P. Xing, Hao Zhang, Joseph~E.
  Gonzalez, and Ion Stoica. 2023.
\newblock \href {https://openreview.net/forum?id=uccHPGDlao} {Judging
  {LLM}-as-a-judge with {MT}-bench and chatbot arena}.
\newblock In \emph{Advances in Neural Information Processing Systems 36
  (NeurIPS 2023) Datasets and Benchmarks Track}.

\bibitem[{Zhong et~al.(2018)Zhong, Xiao, Guo, Tu, Liu, Sun, Feng, Han, Hu,
  Wang, and Xu}]{zhong2018overview}
Haoxi Zhong, Chaojun Xiao, Zhipeng Guo, Cunchao Tu, Zhiyuan Liu, Maosong Sun,
  Yansong Feng, Xianpei Han, Zhen Hu, Heng Wang, and Jianfeng Xu. 2018.
\newblock \href {https://arxiv.org/abs/1810.05851} {Overview of {CAIL2018}:
  Legal judgment prediction competition}.
\newblock \emph{Preprint}, arXiv:1810.05851.

\bibitem[{Zhou et~al.(2024)Zhou, Zhu, Mathur, Zhang, Yu, Qi, Morency, Bisk,
  Fried, Neubig, and Sap}]{zhou2024sotopia}
Xuhui Zhou, Hao Zhu, Leena Mathur, Ruohong Zhang, Haofei Yu, Zhengyang Qi,
  Louis-Philippe Morency, Yonatan Bisk, Daniel Fried, Graham Neubig, and
  Maarten Sap. 2024.
\newblock \href {https://openreview.net/forum?id=mM7VurbA4r} {{SOTOPIA}:
  Interactive evaluation for social intelligence in language agents}.
\newblock In \emph{The Twelfth International Conference on Learning
  Representations (ICLR)}.

\end{thebibliography}

\clearpage
\appendix

\section*{Content of Appendix}
\addcontentsline{toc}{section}{Content of Appendix}

\noindent The appendix is organized into seven parts; each part regroups previously scattered material and adds the supplementary detail referenced from the main text.

\begin{itemize}
  \setlength{\itemsep}{1pt}
  \item[\textbf{A}] \textbf{Role and Persona Setting Details} (\S\ref{app:role-persona}). Role profiles for lawyer, client, and judge agents; visible-state $\times$ stage $\times$ role matrix; LCPF dimensions, level definitions, and level-redistribution policy.
  \item[\textbf{B}] \textbf{Dataset Construction and Additional Statistics} (\S\ref{app:dataset-construction-details}). Source, deduplication, and first/second-instance pairing pipeline; LLM-based field extraction and quality control; Full/Medium/Light split rules; supplementary distribution statistics.
  \item[\textbf{C}] \textbf{Evaluation Metrics in Detail (LongJud-Bench)} (\S\ref{app:longjud-bench-details}). Per-item scoring formulas, metric definitions, normalization rules, the scoring-item-to-capability mapping, 0--10 rubric anchors, and rule-based judicial output alignment.
  \item[\textbf{D}] \textbf{Implementation Details} (\S\ref{app:impl}). Model versions and inference parameters; memory, Skill, and Tool runtime; anonymized role-memory examples; evaluation pipeline and parsing failure handling; compute and token cost; full Tool/Skill catalogue and Skill-card fields.
  \item[\textbf{E}] \textbf{Additional Experiment Results} (\S\ref{app:extra-results}). LCPF persona validation; cross-stage causal dependence; and a per-stage cross-model view.
  \item[\textbf{F}] \textbf{Human Evaluation} (\S\ref{app:human-eval}). Evaluator recruitment and background; task design and assignment plan; scoring rubric and protocol; interface screenshot; informed-consent and data-use statement; human--LLM agreement breakdown.
  \item[\textbf{G}] \textbf{Prompt Templates} (\S\ref{app:prompts}). Bilingual prompt-box figures for the production role prompts, LongJud-Bench benchmark scoring prompts, persona-validation scorer prompt, and experimental LLM-as-Judge prompts.
\end{itemize}

\section{Role and Persona Setting Details}
\label{app:role-persona}
\label{app:scenario-state}

\subsection{Agent Roles and Stage Bindings}

Each case is represented both as a complete life-cycle trajectory and as a set of stage-level tasks. The main text gives the compact stage protocol; the details below preserve the role profile, visible-state rule, and permitted-action description behind that compressed presentation.

\paragraph{Lawyer agents.}
Lawyer agents are the core professional actors across the full litigation cycle, responsible for case analysis, document drafting, and adversarial advocacy. During consultation, they analyze case facts and answer legal questions; during document drafting, they collect information through multi-round dialogue and draft standardized legal documents; during trial, they participate in evidence presentation, cross-examination, and court debate as attorneys. The lawyer role is instantiated on both sides of the adversarial case as plaintiff lawyer $a_{lp}$ and defendant lawyer $a_{ld}$. One of them is the target lawyer agent under evaluation $a^\ast$, while the other serves as the opposing lawyer. The lawyer interface exposes professional case materials, client communications, prior legal artifacts visible to the current stage, and stage-specific Skill/Tool entries.

\paragraph{Client agents.}
Client agents represent ordinary litigation parties whose narratives, goals, and procedural understanding shape the case trajectory. They are instantiated as plaintiff client $a_p$ and defendant client $a_d$. Their interfaces expose party-side facts, consultation questions, procedural progress, and the legal documents or trial events that an ordinary party would observe. The Legal Client Persona Framework conditions how the client discloses facts, asks questions, reacts to litigation risk, and narrates evidence.

\paragraph{Judge agents.}
Judge agents provide procedural control and generate judicial outputs at the two trial stages. They are instantiated as first-instance judge $a_{j1}$ and second-instance judge $a_{j2}$. Their interfaces expose the case record and procedural materials available to the corresponding trial stage, along with ordered court-control actions. The judge role is separated from lawyer and client roles so that trial procedure, evidentiary questioning, and judgment generation are produced through a distinct procedural interface.

\subsection{Stage-Level Procedural Templates}

\paragraph{Legal Consultation.}
Legal Consultation (LC) is the stage where the client describes the dispute and legal concerns, while the target lawyer asks follow-up questions and provides initial legal analysis before formal litigation artifacts are produced. It constructs the initial lawyer-client interaction from $S_c^{(0)}$. The output $O_c^{(1)}$ contains the consultation record and lawyer response, while $H_c^{(1)}$ records the full dialogue trace.

\paragraph{Pre-Trial Document Drafting.}
Pre-Trial Document Drafting transforms collected facts, claims, defenses, and evidence into the first formal pleading or response document. Complaint Drafting (CD) is triggered when the target lawyer represents the plaintiff; Defense Drafting (DD) is triggered when the target lawyer represents the defendant. The output $O_c^{(2)}$ contains the generated complaint or defense documents and the structured drafting record.

\paragraph{First-Instance Trial.}
First-Instance Trial (FIT) constructs the first trial stage from $S_c^{(2)}$ with five participating roles: plaintiff client, defendant client, plaintiff lawyer, defendant lawyer, and first-instance judge. The stage covers opening, court investigation, evidence presentation and cross-examination, judge questioning, court debate, final statements, mediation inquiry, and pronouncement. The judge generates the civil first-instance judgment artifact $O_{\mathrm{FIT}}$ unless mediation is accepted by both parties.

\paragraph{Appeal Determination and Pre-Appellate Drafting.}
Appeal Determination (AD-Det) reads the appeal information after the first-instance judgment and assigns each party to the appellant or appellee role for the appellate stage. Pre-Appellate Document Drafting then transforms the first-instance judgment, appeal requests, appeal reasons, and supplementary materials into written appellate positions. Appeal Drafting (AD) is triggered when the target lawyer represents the appellant; Appeal Response (AR) is triggered when the target lawyer represents the appellee. The output $O_c^{(4)}$ contains second-instance document artifacts and drafting traces.

\paragraph{Second-Instance Trial.}
Second-Instance Trial (SIT) is the final courtroom interaction under the control of the second-instance judge, who reviews the dispute and produces the final judgment. It follows the same structured trial procedure as FIT, with role titles adapted to the second-instance context. The second-instance judgment $J_{\mathrm{final}}\robusteq O_c^{(5)}$ marks the end of the life-cycle simulation.

\begin{algorithm}[!t]
\small
\caption{Structured Civil Trial Procedure. The pseudocode shows how the judge-controlled phase loop collects role responses, updates dispute focus, handles mediation, and returns the next case state.}
\label{alg:structured-trial}
\begin{algorithmic}[1]
\Require Previous case state $S_c^{(t-1)}$, trial type $u\in\{\mathrm{FIT},\mathrm{SIT}\}$, and participating agents $\mathcal{A}_c^{(t)}$
\Ensure Updated case state $S_c^{(t)}$, trial artifact $O_c^{(t)}$, and trace $H_c^{(t)}$
\State $H_c^{(t)}\gets\emptyset$; $Q\gets\mathrm{ExtractDisputes}(S_c^{(t-1)},u)$
\State $\Pi_u\gets\mathrm{OrderedTrialPhases}(u)$
\For{phase $p$ in $\Pi_u$}
    \State $I_p\gets\mathrm{JudgeControl}(p,S_c^{(t-1)},Q,H_c^{(t)})$
    \For{speaker $a$ in $\mathrm{Speakers}(p,u,\mathcal{A}_c^{(t)})$}
        \State $I_{a,c}^{(t,p)}\gets\mathrm{StageInterface}(a,p,S_c^{(t-1)},I_p,Q)$
        \State $r_a\gets\mathrm{AgentAct}(I_{a,c}^{(t,p)},H_c^{(t)})$
        \State $H_c^{(t)}\gets H_c^{(t)}\cup\{(p,a,r_a)\}$
    \EndFor
    \State $Q\gets\mathrm{UpdateDisputes}(Q,p,H_c^{(t)})$
    \If{$p\robusteq\mathrm{Mediation}$ and $\mathrm{AcceptBothParties}(H_c^{(t)})$}
        \State $O_c^{(t)}\gets\mathrm{BuildMediationRecord}(H_c^{(t)},Q)$
        \State \textbf{break}
    \EndIf
\EndFor
\If{$O_c^{(t)}$ is not assigned}
    \State $O_c^{(t)}\gets\mathrm{DeliberateAndJudge}(S_c^{(t-1)},Q,H_c^{(t)})$
\EndIf
\State $M_c^{(t)}\gets\mathrm{MemUpdate}(M_c^{(t-1)},O_c^{(t)},H_c^{(t)})$
\State $S_c^{(t)}\gets\mathrm{Transition}(S_c^{(t-1)},O_c^{(t)},H_c^{(t)},M_c^{(t)})$
\State \Return $(S_c^{(t)},O_c^{(t)},H_c^{(t)})$
\end{algorithmic}
\end{algorithm}

\subsection{\texorpdfstring{Visible-State $\times$ Stage $\times$ Role Matrix}{Visible-State x Stage x Role Matrix}}
\label{app:visible-state-matrix}

Table~\ref{tab:visible-state-matrix} records which case fields each agent sees at each stage. The matrix is derived from the visible-state rule $V_{a,c}^{(t)}$ in Equation~\ref{eq:stage-interface}; a checkmark means the field is included in the agent's stage interface, and a blank cell means the field is filtered out before prompt assembly. Reference answers, hidden judgment fields, and the opposing party's private memory never appear in any agent's $V_{a,c}^{(t)}$ at any stage.

\begin{table*}[!t]
\centering
\footnotesize
\setlength{\tabcolsep}{3pt}
\renewcommand{\arraystretch}{0.95}
\begin{tabularx}{\textwidth}{>{\raggedright\arraybackslash}p{0.27\textwidth}cccccc Y}
\toprule
Visible field & LC & CD/DD & FIT & AD/AR & SIT & Judge (FIT/SIT) & Filter note \\
\midrule
Party-side facts (own side) & C,L & C,L & C,L & C,L & C,L & J & Opponent side not exposed to either client. \\
Known evidence (own side) & C,L & C,L & C,L & C,L & C,L & J & Opponent evidence revealed only after court investigation. \\
Litigation goal / bottom line (own side) & C,L & C,L & C,L & C,L & C,L & --- & Judge does not read litigation goals. \\
Consultation questions \& reference Q list & C & --- & --- & --- & --- & --- & Reference \emph{answers} are evaluation-only. \\
Plaintiff claims / appeal requests & --- & C,L & C,L,J & C,L & C,L,J & J & Surfaced to opposing side at CD/DD or AD/AR via document. \\
First-instance pleadings ($O_c^{(2)}$) & --- & own & C,L,J & C,L & C,L,J & J & Drafted by own lawyer; opponent reads it at FIT. \\
First-instance judgment ($O_{\mathrm{FIT}}$) & --- & --- & --- & C,L & C,L,J & J & Generated at FIT; reused as a reference at AD/AR and SIT. \\
Appellate pleadings ($O_c^{(4)}$) & --- & --- & --- & own & C,L,J & J & Drafted by own lawyer; opponent reads at SIT. \\
Lawyer global memory $G_{a_l,c}^{(t)}$ & L & L & L & L & L & --- & Each lawyer reads only their own memory; opponent's memory hidden. \\
Client global memory $G_{a_c,c}^{(t)}$ & C & C & C & C & C & --- & Each client reads only their own memory. \\
In-scenario local memory $L_{a,c}^{(t)}$ & all & all & all & all & all & J & Limited to the current scenario's dialogue trace. \\
Reference answers / hidden judgment refs & --- & --- & --- & --- & --- & --- & Evaluation-only; gated by \texttt{evaluation\_only=true}. \\
\bottomrule
\end{tabularx}
\caption{Visible-state $\times$ stage $\times$ role matrix. ``C'' indicates the client of the corresponding side, ``L'' the lawyer of the corresponding side, ``J'' the presiding judge of the trial stage, and ``---'' means the field is not exposed at that stage. ``own'' means only the side that drafted the artifact sees it before opposing exposure. The matrix is enforced by the visible-state rule $V_{a,c}^{(t)}$ in Equation~\ref{eq:stage-interface}; fields outside the rule are stripped before prompt assembly.}
\label{tab:visible-state-matrix}
\end{table*}

\subsection{Legal Client Persona Framework}
\label{app:dataset-construction-details}

The Legal Client Persona Framework (LCPF) defines four legal-scene dimensions. Table~\ref{tab:lcpf-dimensions} states the behavioral meaning of each dimension. Each dimension is assigned a high, medium, or low level. The medium-level redistribution rule applies only when the LLM-based persona generator returns more than 60\% medium for any dimension on a given Light/Medium split: in that case the medium label of the over-represented cases is reassigned to high or low uniformly at random, conditioned on case-cause balance, until the medium share falls below 60\%. This rule avoids overly homogeneous client behavior without changing the semantic definition of the dimensions; the per-dimension level distribution before and after redistribution is logged with the case seed for reproducibility.

\begin{table*}[!t]
\centering
\footnotesize
\setlength{\tabcolsep}{2.4pt}
\renewcommand{\arraystretch}{0.94}
\begin{tabularx}{\textwidth}{>{\raggedright\arraybackslash}p{0.17\textwidth}>{\raggedright\arraybackslash}p{0.25\textwidth}>{\raggedright\arraybackslash}p{0.24\textwidth}Y}
\toprule
Dimension & High & Medium & Low \\
\midrule
Legal Literacy & Understands basic procedural rights, evidentiary burdens, and the distinction between facts and legal claims. & Understands common legal terms but needs guidance on procedural consequences. & Confuses legal concepts and often expresses claims as everyday grievances. \\
Information Disclosure Willingness & Proactively discloses favorable and unfavorable facts relevant to the dispute. & Answers direct questions but may omit uncertain or embarrassing details. & Withholds unfavorable information or gives incomplete answers until pressed. \\
Emotional Stability & Communicates calmly and can follow repeated legal guidance. & Shows stress but remains responsive to lawyer guidance. & Becomes anxious, angry, or distracted under procedural pressure. \\
Narrative Proficiency & Presents chronology, actors, evidence, and disputed points in an organized way. & Provides usable facts but needs help ordering them. & Provides fragmented narratives with missing chronology or unclear evidence links. \\
\bottomrule
\end{tabularx}
\caption{Legal Client Persona dimensions and level meanings. Each row defines one persona dimension and contrasts the behavioral expectations for high, medium, and low levels during legal interaction.}
\label{tab:lcpf-dimensions}
\end{table*}

\section{Dataset Construction and Additional Statistics}
\label{app:dataset}

\subsection{Source Collection and Pair Construction}

The source corpus is collected from China Judgments Online (\textit{wenshu.court.gov.cn}). The crawl pulls all civil first-instance and civil second-instance judgments published in the configured window, retaining the rendered text body and the case number. Duplicate filings are removed by a key composed of (court name, full case number, judgment date, hash of party-name set); when collisions remain, the longer judgment text is kept.

First/second-instance pairing is then performed within each court hierarchy. For each second-instance judgment, the pairing routine searches the first-instance pool for a candidate sharing the same lower-court case number cited inside the appellate text. A candidate is retained only when the normalized party-name set matches exactly, the cause of action matches exactly, and the judgment-date order is respected (first instance precedes second instance). We further remove second-instance records that did not proceed to a substantive appellate hearing, including cases resolved only through withdrawal, non-acceptance, procedural dismissal, or other non-hearing dispositions. Pairs that fail any of these checks are discarded. After pairing, 75{,}309 (first, second) tuples remain.

\subsection{Field Extraction and Quality Control}

Structured fields are extracted from each raw judgment by an LLM-based extractor (DeepSeek-V3.2)~\citep{deepseek2025v32} using a stage-typed schema covering party identifiers, claims and defenses, facts and reasons, evidence list, court findings, legal references, judgment disposition, and the analogous appellate fields. The extractor is prompted to write \texttt{null} when a field is not present in the source text and is forbidden from inferring missing identifiers. Extraction outputs are validated by a JSON-schema checker; failures are re-tried up to three times before the case is dropped.

\subsection{Splits}

Three splits are derived from the paired corpus. \textbf{Full} retains all 75{,}309 pairs across over 500 causes of action. \textbf{Medium} samples 1{,}000 cases stratified by cause-of-action---the top 100 most frequent causes contribute 10 cases each, sampled without replacement under a fixed seed (\texttt{seed=20251217}). \textbf{Light} subsamples Medium down to 100 cases by retaining the top 20 most frequent causes with five cases each, again under a fixed seed. The splits are stored alongside per-case case-IDs so any sampling can be reproduced.

\begin{table}[!t]
\centering
\scriptsize
\setlength{\tabcolsep}{3pt}
\renewcommand{\arraystretch}{0.96}
\begin{tabularx}{\columnwidth}{lrrY}
\toprule
Split & Cases & Causes & Sampling rule \\
\midrule
Full & 75{,}309 & 500+ & All paired cases after matching and filtering. \\
Medium & 1{,}000 & 100 & Top 100 causes, 10 cases each. \\
Light & 100 & 20 & Top 20 causes, five cases each. \\
\bottomrule
\end{tabularx}
\caption{Dataset split sizes used by LongJud-Bench. Full retains the complete paired corpus after matching and filtering; Medium and Light are deterministic cause-balanced subsets sampled with the fixed seed \texttt{20251217}.}
\label{tab:dataset-splits}
\end{table}

\subsection{Field Groups and Additional Statistics}

Table~\ref{tab:dataset-field-groups} summarizes the major field groups used to construct runnable cases. Figure~\ref{fig:dataset-composition} reports the court-level and top-category cause distribution of the Full split; the broad mix shown there motivates the cause-balanced sampling used for Medium and Light.

\begin{table*}[!t]
\centering
\footnotesize
\setlength{\tabcolsep}{3pt}
\renewcommand{\arraystretch}{0.94}
\begin{tabularx}{\textwidth}{>{\raggedright\arraybackslash}p{0.20\textwidth}>{\raggedright\arraybackslash}p{0.27\textwidth}Y}
\toprule
Field group & Representative fields & Use in the life-cycle environment \\
\midrule
Case metadata & Cause of action, court level, court name, procedural status, judgment date & Defines the procedural setting and organizes cases by legal domain. \\
Party information & Party role, natural-person or organization type, residence or registered address, representative information when available & Initializes litigant roles and document-party fields. \\
First-instance procedure & Claims, facts and reasons, evidence list, defense opinions, court findings, legal references, judgment disposition & Provides reference materials for first-instance drafting, trial, and judgment alignment. \\
Second-instance procedure & Appeal requests, appeal reasons, appellee defenses, new evidence, appellate findings, affected first-instance items, final disposition & Supports appeal-role mapping, appellate drafting, and second-instance trial evaluation. \\
Legal Persona & Legal Literacy, Information Disclosure Willingness, Emotional Stability, Narrative Proficiency & Conditions client behavior in consultation, drafting, and trial interaction. \\
Consultation supervision & Party-side legal questions and reference answers & Provides question-level references for LC evaluation. \\
\bottomrule
\end{tabularx}
\caption{Major field groups in the structured case seed. The table maps each extracted or generated field group to the information it supplies when constructing and running a life-cycle civil case.}
\label{tab:dataset-field-groups}
\end{table*}

Each reference answer is produced by a separate LLM call (DeepSeek-V3.2) that takes the case's accepted facts and applicable statutes as input. The result is used only as an evaluation reference for LC scoring and is not provided to agents during simulation.

\subsection{License and Terms of Use}

The judgment documents are collected from China Judgments Online (wenshu.court.gov.cn), which publishes civil judgments under public access. We use the data strictly for non-commercial academic research.

\paragraph{Created artifacts.}
We will release LegalWorld code under the MIT License and LongJud-Bench under CC BY-NC 4.0 to support academic use while restricting commercial deployment in real legal services.

\paragraph{Used artifacts.}
The LLM backbones (Claude, Qwen, Kimi, GPT, DeepSeek, GLM) are accessed via their respective official APIs, and their use in this paper follows each provider's published terms of service.

\section{Evaluation Metrics in Detail (LongJud-Bench)}
\label{app:longjud-bench-details}

\subsection{Stage Subitems and Scoring Methods}

LongJud-Bench scores each underlying item and maps the normalized outputs into the eight capabilities reported in Table~\ref{tab:cross-model}. The retained per-item outputs allow inspection of which consultation, drafting, or trial item contributed to each capability. Every item is normalized to $[0,1]$: exact-match metrics return $0$ or $1$ after field normalization, while LLM-as-Judge metrics use a 0--10 rubric and are divided by 10. Table~\ref{tab:evaluation-subitems} summarizes the scoring items, the capabilities they feed, their metric units, and their reference sources.

\begin{table*}[!t]
\centering
\scriptsize
\setlength{\tabcolsep}{2pt}
\renewcommand{\arraystretch}{0.96}
\begin{tabularx}{\textwidth}{>{\raggedright\arraybackslash}p{0.08\textwidth}>{\raggedright\arraybackslash}p{0.20\textwidth}>{\raggedright\arraybackslash}p{0.24\textwidth}>{\raggedright\arraybackslash}p{0.17\textwidth}Y}
\toprule
Stage & Capabilities & Main scoring items & Metric unit & Reference source \\
\midrule
LC & Issue spotting & Legal relationship identification, applicable rules/statutes, risk explanation, procedural advice and actionability & Per-question LLM-as-Judge score, 0--10 & Generated reference answers grounded in case facts and statutes. \\
CD/DD & Party identification; claim construction; fact marshalling; evidence marshalling & Party identity and procedural slots; claims or defenses; facts and reasons; evidence list & Exact match for identity/procedural slots; 0--10 semantic scoring for narrative/legal slots & First-instance structured judgment fields and party-side records. \\
FIT & Position consistency; evidentiary advocacy; legal reasoning & Consistency between the statements and the pleaded claims/defenses; fact-and-evidence use; legal-reasoning sufficiency & Per-trial-phase target-lawyer statements, each dimension 0--10 & First-instance case record, pleadings, evidence, and legal standards. \\
AD/AR & Party identification; claim construction; fact marshalling; evidence marshalling & Appellate role and party slots; appeal requests or responses; appeal reasons; new evidence & Exact match for structured slots; 0--10 semantic scoring for appellate arguments & Second-instance structured judgment fields and appellate records. \\
SIT & Position consistency; evidentiary advocacy; legal reasoning & Consistency between the statements and the appeal/response; fact-and-new-evidence use; legal-reasoning sufficiency & Per-appellate-phase target-lawyer statements, each dimension 0--10 & Appellate pleadings, first-instance judgment, new evidence, and final judgment reference. \\
\bottomrule
\end{tabularx}
\caption{LongJud-Bench scoring items and the capabilities they feed in Table~\ref{tab:cross-model}. Drafting and trial items on the first-instance side (CD/DD, FIT) and the second-instance side (AD/AR, SIT) populate the same capabilities, reported as the two halves of each cell. Exact-match slots are scored as binary normalized fields; semantic and trial items use 0--10 LLM-as-Judge rubrics before normalization.}
\label{tab:evaluation-subitems}
\end{table*}

\subsection{Stage Formulas}

The formulas below are written per stage (LC, CD/DD, FIT, AD/AR, SIT) because the underlying evidence is collected at those procedural points. Each formula specifies how the Bench scores one procedural part, and the resulting item scores feed the capabilities exactly as listed in Table~\ref{tab:evaluation-subitems} (e.g., the FIT/SIT trial dimensions below are the items behind the position-consistency, evidentiary-advocacy, and legal-reasoning capabilities).

Let $\operatorname{norm}(x)$ denote the metric normalization function. For exact-match fields, $\operatorname{norm}(x)=x$ with $x\in\{0,1\}$. For an LLM-as-Judge rating $r\in[0,10]$, $\operatorname{norm}(r)=r/10$. Each stage score below is therefore in $[0,1]$.

In LC, the client raises $n$ legal questions, each paired with a reference answer generated from case facts and applicable statutes. Each question is evaluated on the dimension set $\mathcal{M}_{\mathrm{LC}}=\{\mathrm{relationship},\mathrm{rule},\mathrm{risk},\mathrm{advice}\}$. The score is:
\begin{equation}
S_{\mathrm{LC}}\robusteq
\frac{1}{n}\sum_{i=1}^{n}
\frac{1}{|\mathcal{M}_{\mathrm{LC}}|}
\sum_{m\in\mathcal{M}_{\mathrm{LC}}}\operatorname{norm}(r_{i,m}).
\end{equation}
Here, $r_{i,m}$ is the 0--10 score for dimension $m$ of consultation question $i$.

For drafting stages, document evaluation integrates exact matching and semantic evaluation. For stage $t\in\{\mathrm{CD/DD},\mathrm{AD/AR}\}$, let $F_t$ be the set of exact-match fields and $G_t$ be the set of semantic fields:
\begin{equation}
S_t\robusteq
\frac{
\sum_{f\in F_t}\operatorname{norm}(e_f)
+\sum_{g\in G_t}\operatorname{norm}(r_g)}
{|F_t|+|G_t|}.
\end{equation}
$F_{\mathrm{CD/DD}}$ covers party identity and procedural slots; $G_{\mathrm{CD/DD}}$ covers claims or defenses, facts and reasons, evidence use, requested disposition, and coherence. $F_{\mathrm{AD/AR}}$ covers appellate role, party, and request/response slots; $G_{\mathrm{AD/AR}}$ covers appeal reasons, new evidence, linkage to the first-instance judgment, and coherence. Only the role-conditional sub-scenario actually taken by the target lawyer is scored.

For FIT and SIT, trial evaluation scores target-lawyer statements by trial phase. Let $P_t$ be the set of scored phases and $\mathcal{D}_t$ the dimension set for stage $t$. For both FIT and SIT, $\mathcal{D}_t$ contains three dimensions---consistency between the statements and the pleaded position, fact-and-evidence use, and legal-reasoning sufficiency---which map to the position-consistency, evidentiary-advocacy, and legal-reasoning capabilities:
\begin{equation}
\begin{aligned}
S_t&\robusteq
\frac{1}{|P_t|}\sum_{p\in P_t}
\left(
\frac{1}{|\mathcal{D}_t|}
\sum_{d\in\mathcal{D}_t}\operatorname{norm}(r_{p,d})
\right),\\
&\hspace{18pt}t\in\{\mathrm{FIT},\mathrm{SIT}\}.
\end{aligned}
\end{equation}
If the target lawyer provides no required statement for a scored phase, the missing phase receives 0 on the affected dimensions. The normalized items are then grouped into the eight capabilities of Table~\ref{tab:cross-model}, which are reported separately so that capability-level trade-offs across backbones stay visible.

\subsection{Evaluation Rubrics}
\label{app:evaluation-rubrics}

The following rubrics make the evaluation anchors explicit. They are used to guide both process-authenticity checks and stage-level scoring; they do not introduce additional experimental claims beyond the main paper.

\begin{table*}[!t]
\centering
\scriptsize
\setlength{\tabcolsep}{2pt}
\renewcommand{\arraystretch}{0.96}
\begin{tabularx}{\textwidth}{>{\raggedright\arraybackslash}p{0.16\textwidth}>{\raggedright\arraybackslash}p{0.25\textwidth}>{\raggedright\arraybackslash}p{0.25\textwidth}Y}
\toprule
Dimension & 9--10 & 7--8 & 5--6 \\
\midrule
Stage authenticity & Procedure is complete or nearly complete, with natural progression and clear role turns. & Procedure is mostly complete, with light omissions or abrupt transitions. & The stage remains interpretable but has limited procedural coverage or weak transitions. \\
Role consistency & Behavior is consistently aligned with role responsibility, stance, and professional identity. & Role identity is mostly stable with occasional generic expressions. & The role remains recognizable but sometimes borrows another role's reasoning style or communicative posture. \\
Trial advocacy & Statements address disputed issues, evidence, legal basis, and procedural position with strong organization. & Statements cover most relevant issues with some evidentiary or legal links left implicit. & Advocacy contains useful points but has limited issue structure or evidence linkage. \\
Document quality & The document is complete, procedurally appropriate, factually grounded, and legally coherent. & The document is usable with light detail gaps, weak transitions, or limited legal elaboration. & The document has recognizable structure but limited factual, evidentiary, or legal completeness. \\
\bottomrule
\end{tabularx}

\vspace{0.35em}
\begin{tabularx}{\textwidth}{>{\raggedright\arraybackslash}p{0.16\textwidth}>{\raggedright\arraybackslash}p{0.39\textwidth}Y}
\toprule
Dimension & 3--4 & 0--2 \\
\midrule
Stage authenticity & Procedural coverage is sparse or ordering is difficult to follow. & The stage provides little reliable procedural signal. \\
Role consistency & Role stance or communication style changes repeatedly. & The role identity provides little reliable signal. \\
Trial advocacy & Statements are sparse and only loosely connected to evidence or law. & Statements provide little usable advocacy signal. \\
Document quality & The document has substantial gaps in required sections or support. & The document provides little usable drafting signal for the intended procedural task. \\
\bottomrule
\end{tabularx}
\caption{Supplementary 0--10 rubric anchors. The two-part table lists qualitative anchors for high, medium, low, and failing score ranges so that process and output scores are interpreted consistently.}
\label{tab:evaluation-rubrics}
\end{table*}

\subsection{Rule-Based Judgment Alignment Metric}
\label{app:rule-alignment}

The rule-based judgment alignment metric evaluates whether generated judgment artifacts align with real judicial outputs on structured legal elements; the alignment scores are reported in the main text (\S\ref{sec:judicial-alignment}, Table~\ref{tab:judicial-output-alignment}). Table~\ref{tab:rule-alignment-dimensions} lists the stage-specific dimensions, and the scoring formula follows below.

\begin{table}[!t]
\centering
\footnotesize
\setlength{\tabcolsep}{2.4pt}
\begin{tabularx}{\columnwidth}{>{\raggedright\arraybackslash}p{0.24\columnwidth}>{\raggedright\arraybackslash}p{0.18\columnwidth}Y}
\toprule
Dimension & Stage & Evaluated elements \\
\midrule
Verdict & FIT/SIT & Judgment disposition and supported or rejected requests. \\
Reasoning & FIT/SIT & Factual findings, dispute focus, and adjudicative rationale. \\
Legal reference & FIT/SIT & Law titles, article references, and provision families. \\
Entity & FIT/SIT & Party identities, role mapping, amounts, and other structured entities. \\
Structure & FIT/SIT & Presence and organization of standard judgment sections. \\
Appeal action & SIT & Whether the appellate judgment affirms, reverses, remands, or modifies affected items. \\
\bottomrule
\end{tabularx}
\caption{Rule-based judgment-alignment dimensions. The table lists the structured judgment elements extracted from generated and real judgments for first- and second-instance alignment scoring.}
\label{tab:rule-alignment-dimensions}
\end{table}

For each set-like dimension, let $G_m$ denote the set extracted from the real judgment and $\hat{G}_m$ denote the set extracted from the generated judgment. Precision, recall, and F1 are:
\begin{align*}
P_m&=\frac{|G_m\cap \hat{G}_m|}{|\hat{G}_m|},\\
R_m&=\frac{|G_m\cap \hat{G}_m|}{|G_m|},\\
F1_m&=\frac{2P_mR_m}{P_m+R_m}.
\end{align*}
When a dimension produces a partial numeric match $x$, the score transformation is:
\[
\phi(x)\robusteq\sqrt{\min(1,\max(0,x))}.
\]
Unavailable components are skipped rather than scored as zero. The reported stage score is:
\[
S_s\robusteq 10\cdot \frac{1}{|\mathcal{M}_s|}\sum_{m\in \mathcal{M}_s} S_m ,
\]
where $\mathcal{M}_s$ is the available metric set for stage $s$.

\section{Implementation Details}
\label{app:impl}

\subsection{Model Versions and Inference Parameters}

The LLM-as-Judge scorer uses Claude-Sonnet-4.6 throughout (\texttt{claude-sonnet-4-6}). The default lawyer-agent and environment-agent backbone is Qwen3.5-Plus (\texttt{qwen3.5-plus}); cross-model experiments evaluate Qwen3.5-Plus itself and also swap the target lawyer backbone among Kimi-K2.5 (\texttt{kimi-k2.5}), GPT-5.2 (\texttt{gpt-5.2}), DeepSeek-V4-Flash (\texttt{deepseek-v4-flash}), GLM-4.7 (\texttt{glm-4.7}), and Qwen3.5-Flash (\texttt{qwen3.5-flash}) while keeping all non-target roles on Qwen3.5-Plus. All calls go through each provider's official API. Inference parameters are held constant: \texttt{temperature}=0.7, \texttt{top\_p}=0.95, per-call \texttt{max\_tokens}=4096, with a per-scenario turn budget of 30 turns for dialogue scenarios (LC, CD/DD, AD/AR) and 60 turns for trial scenarios (FIT, SIT).

\subsection{Memory, Skill, and Tool Runtime}

Global case memory is stored as a JSON object whose top-level keys are \texttt{facts}, \texttt{evidence}, \texttt{claims}, \texttt{defenses}, \texttt{procedural\_progress}, \texttt{client\_profile}, \texttt{positions}, and \texttt{notes}; writes go through the bounded \texttt{revise} and \texttt{expand} operations described in Section~\ref{sec:cross-stage-memory}. Tool calls are throttled at the step level: each agent may issue at most eight Tool calls during a single simulation step. Calls that exceed this step-level cap are rejected and surfaced to the agent as an explicit failure message rather than silently dropped. Retries follow a fixed three-attempt schedule on JSON-decode failures, transient HTTP errors, and rate-limit responses, with exponential backoff between attempts. Skill instructions are injected on the first turn a Skill is required and stay in the prompt for the rest of the scenario.

\begin{figure*}[!t]
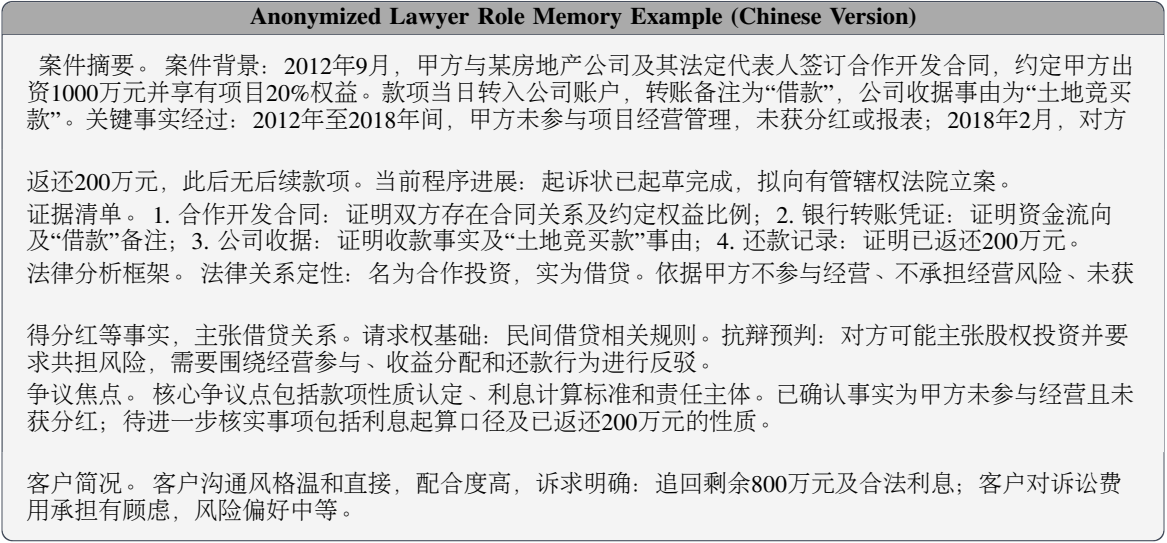

\centering
\begin{minipage}{0.96\textwidth}
\begin{promptbox}{Anonymized Lawyer Role Memory Example (Chinese Version)}
\begin{CJK*}{UTF8}{gbsn}
\textbf{案件摘要。} 案件背景：2012年9月，甲方与某房地产公司及其法定代表人签订合作开发合同，约定甲方出资1000万元并享有项目20\%权益。款项当日转入公司账户，转账备注为“借款”，公司收据事由为“土地竞买款”。关键事实经过：2012年至2018年间，甲方未参与项目经营管理，未获分红或报表；2018年2月，对方返还200万元，此后无后续款项。当前程序进展：起诉状已起草完成，拟向有管辖权法院立案。

\textbf{证据清单。} 1. 合作开发合同：证明双方存在合同关系及约定权益比例；2. 银行转账凭证：证明资金流向及“借款”备注；3. 公司收据：证明收款事实及“土地竞买款”事由；4. 还款记录：证明已返还200万元。

\textbf{法律分析框架。} 法律关系定性：名为合作投资，实为借贷。依据甲方不参与经营、不承担经营风险、未获得分红等事实，主张借贷关系。请求权基础：民间借贷相关规则。抗辩预判：对方可能主张股权投资并要求共担风险，需要围绕经营参与、收益分配和还款行为进行反驳。

\textbf{争议焦点。} 核心争议点包括款项性质认定、利息计算标准和责任主体。已确认事实为甲方未参与经营且未获分红；待进一步核实事项包括利息起算口径及已返还200万元的性质。

\textbf{客户简况。} 客户沟通风格温和直接，配合度高，诉求明确：追回剩余800万元及合法利息；客户对诉讼费用承担有顾虑，风险偏好中等。
\end{CJK*}
\end{promptbox}
\caption{An anonymized lawyer role-memory excerpt (\textit{Chinese Version}). The lawyer memory keeps a professional case record that separates facts, evidence, legal analysis, dispute focuses, and client-brief information.}
\label{fig:lawyer-role-memory-cn}
\end{minipage}
\end{figure*}

Figures~\ref{fig:lawyer-role-memory-cn}--\ref{fig:client-role-memory-en} give bilingual anonymized excerpts of the role-specific structured memory records produced by the runtime memory writers. The examples preserve the field structure used by the system while omitting party-identifying names.

\begin{figure*}[!t]
\centering
\begin{minipage}{0.96\textwidth}
\begin{promptbox}{Anonymized Lawyer Role Memory Example (English Version)}
\textbf{Case Summary.} Case background: in September 2012, Party A entered into a cooperative real-estate development agreement with a real-estate company and its legal representative. Party A contributed RMB 10 million and was to hold a 20\% project interest. The funds were transferred to the company account on the same day; the transfer note described the payment as a loan, while the company receipt described the purpose as land-bidding funds. Key factual sequence: from 2012 to 2018, Party A did not participate in project operation or management and received no dividends or financial reports. In February 2018, the counterparty returned RMB 2 million, with no further payments afterward. Current procedural progress: the civil complaint has been drafted and is ready for filing with a court of competent jurisdiction.

\textbf{Evidence Ledger.} 1. Cooperative development agreement: proves the contractual relationship and agreed interest ratio; 2. bank-transfer voucher: proves fund flow and the loan notation; 3. company receipt: proves receipt of funds and the stated land-bidding purpose; 4. repayment record: proves the RMB 2 million partial repayment.

\textbf{Legal Framework.} Legal characterization: nominally a cooperative investment, substantively a loan. Because Party A did not participate in operation, did not bear business risk, and did not receive dividends, the claim frames the relationship as private lending. Basis of claim: private-lending rules. Anticipated defense: the counterparty may characterize the arrangement as equity investment with shared risk, so the response should focus on lack of operational participation, lack of profit distribution, and the repayment conduct.

\textbf{Dispute Focus.} Core disputes include characterization of the payment, interest-calculation standard, and liable party. Confirmed facts include Party A's lack of operational participation and lack of dividends. Items requiring further confirmation include the interest start date and the characterization of the RMB 2 million repayment.

\textbf{Client Brief.} The client communicates directly and calmly, cooperates well, and has a clear demand: recover the remaining RMB 8 million and lawful interest. The client is concerned about litigation costs and has a medium risk preference.
\end{promptbox}
\caption{An anonymized lawyer role-memory excerpt (\textit{English Version}). The lawyer memory keeps a professional case record that separates facts, evidence, legal analysis, dispute focuses, and client-brief information.}
\label{fig:lawyer-role-memory-en}
\end{minipage}
\end{figure*}

\begin{figure*}[!t]
\centering
\begin{minipage}{0.96\textwidth}
\begin{promptbox}{Anonymized Client Role Memory Example (Chinese Version)}
\begin{CJK*}{UTF8}{gbsn}
\textbf{案件进展。} 已完成法律咨询，并明确基本诉讼策略：将相关个人与公司列为共同被告；确认利息主张采用合法利率上限口径；决定启动起诉程序，下一步准备起草起诉状。已完成起诉状初稿核对，确认本金主张为800万元，利息按约定口径计算；为提高立案稳定性与胜诉把握，暂不主张违约金及住宿费、保全费、律师费等维权成本；待补充原告住址并核实利息起算日期后即可定稿立案。

\textbf{核心诉求。} 核心诉求是追回剩余本金800万元及合法利息。客户接受暂不主张违约金和部分维权成本的稳妥方案，倾向于尽快立案诉讼；若对方能够一次性支付主要款项，客户可以考虑调解。
\end{CJK*}
\end{promptbox}
\caption{An anonymized client role-memory excerpt (\textit{Chinese Version}). The client memory preserves the party-side understanding of procedural progress, litigation goals, concessions, and settlement bottom line.}
\label{fig:client-role-memory-cn}
\end{minipage}
\end{figure*}

\begin{figure*}[!t]
\centering
\begin{minipage}{0.96\textwidth}
\begin{promptbox}{Anonymized Client Role Memory Example (English Version)}
\textbf{Case Progress.} The legal consultation has been completed and the basic litigation strategy has been clarified: list the relevant individual and company as co-defendants; adopt a lawful upper-bound interest standard; initiate litigation, with the next step being preparation of the civil complaint. The draft complaint has been reviewed; the principal claim is confirmed as RMB 8 million, with interest calculated under the agreed lawful standard. To improve filing stability and litigation prospects, the client will temporarily refrain from claiming liquidated damages and certain enforcement costs such as lodging, preservation, and attorney fees. After supplementing the plaintiff's address and confirming the interest start date, the document can be finalized for filing.

\textbf{Core Demands.} The core demand is to recover the remaining RMB 8 million principal and lawful interest. The client accepts the conservative plan of temporarily not claiming liquidated damages and some rights-protection costs, prefers filing promptly, and may consider settlement if the counterparty can pay the main amount in one lump sum.
\end{promptbox}
\caption{An anonymized client role-memory excerpt (\textit{English Version}). The client memory preserves the party-side understanding of procedural progress, litigation goals, concessions, and settlement bottom line.}
\label{fig:client-role-memory-en}
\end{minipage}
\end{figure*}

\subsection{Evaluation Pipeline}

Each completed case is processed by the evaluation runner, which (i) loads the persisted scenario outputs $O_c^{(\leq 5)}$ and dialogue traces $H_c^{(\leq 5)}$, (ii) calls the LLM-as-Judge with the stage-specific rubric, agent output, and evaluation reference fields, and (iii) parses the structured score response. Parsing failures fall back to a one-shot regeneration with a stricter ``return JSON only'' system prompt; cases whose regeneration also fails are flagged for manual audit and excluded from the aggregate score for that condition.

\subsection{Compute and Token Cost}

A complete life-cycle run for one case averages 500{,}000 tokens (prompt + completion), distributed approximately as 6\% LC, 17\% CD/DD, 35\% FIT, 12\% AD/AR, 30\% SIT. Running one lawyer backbone over the Light split consumes tens of millions of tokens and several wall-clock hours under the default 20-way batch concurrency, and the LLM-as-Judge sweep over the same split is lighter. The cross-model experiments in Section~\ref{sec:cross-model} together consume on the order of 0.3B tokens.

\begin{table*}[!t]
\centering
\scriptsize
\setlength{\tabcolsep}{2.1pt}
\renewcommand{\arraystretch}{0.92}
\begin{tabularx}{\textwidth}{>{\raggedright\arraybackslash}p{0.18\textwidth}>{\raggedright\arraybackslash}p{0.07\textwidth}>{\raggedright\arraybackslash}p{0.13\textwidth}>{\raggedright\arraybackslash}p{0.14\textwidth}Y}
\toprule
Component & Type & Category & Roles / Stages & Function \\
\midrule
Client memory writing & Skill & Memory & Client / all stages & Guides clients to preserve stable facts, litigation goals, and perceived case progress. \\
Lawyer memory writing & Skill & Memory & Lawyer / all stages & Guides lawyers to update facts, evidence ledger, legal analysis, dispute focus, client profile, and strategy fields. \\
Complaint drafting & Skill & Document drafting & Plaintiff lawyer / CD & Structures plaintiff information, claims, facts, reasons, and evidence for civil complaint drafting. \\
Defense drafting & Skill & Document drafting & Defendant lawyer / DD & Structures defense opinions, factual rebuttals, evidence, and procedural responses for civil defense drafting. \\
Appeal drafting & Skill & Document drafting & Appellant lawyer / AD & Organizes appeal requests, reasons, challenges to the first-instance judgment, and new evidence. \\
Appeal response drafting & Skill & Document drafting & Appellee lawyer / AR & Organizes responses to appeal requests, defense opinions, and supplementary evidence in second instance. \\
\midrule
Skill provider & Tool & Runtime supply & All roles / all stages & Supplies the stage-appropriate Skill instructions to the agent context. \\
Statute retrieval & Tool & Legal retrieval & Lawyer, judge / all stages & Retrieves relevant statutes and legal provisions for legal relationship analysis and reasoning. \\
Prior-artifact reader & Tool & Artifact access & Lawyer, evaluator & Reads earlier stage outputs and evaluation-facing artifacts within the current case boundary. \\
Client memory reader & Tool & Memory & Client / all stages & Reads the client's structured case memory. \\
Client memory writer & Tool & Memory & Client / all stages & Writes updated client memory under field-level constraints. \\
Lawyer memory reader & Tool & Memory & Lawyer / all stages & Reads the lawyer's structured professional case memory. \\
Lawyer memory writer & Tool & Memory & Lawyer / all stages & Writes updated lawyer memory under field-level constraints. \\
Complaint exporter & Tool & Document generation & Plaintiff lawyer / CD & Converts the completed complaint text into the standardized document artifact. \\
Defense exporter & Tool & Document generation & Defendant lawyer / DD & Converts the completed defense text into the standardized document artifact. \\
Appeal exporter & Tool & Document generation & Appellant lawyer / AD & Converts the completed appeal text into the standardized document artifact. \\
Appeal response exporter & Tool & Document generation & Appellee lawyer / AR & Converts the completed appeal response into the standardized document artifact. \\
First-instance judgment exporter & Tool & Judgment generation & Judge / FIT & Converts the first-instance judgment into the standardized judgment artifact. \\
Second-instance judgment exporter & Tool & Judgment generation & Judge / SIT & Converts the second-instance judgment into the standardized final judgment artifact. \\
Case retrieval & Tool & Legal retrieval & Lawyer, judge & Retrieves similar cases and adjudicative references for legal reasoning. \\
Citation checker & Tool & Document review & Lawyer, judge & Checks whether cited statutes exist and whether article references are consistent. \\
Document comparator & Tool & Document review & Lawyer, evaluator & Compares legal documents and highlights differences in claims, evidence, and dispute focuses. \\
Evaluation runner & Tool & Evaluation & Evaluator & Runs stage-level and life-cycle LongJud-Bench evaluation. \\
\bottomrule
\end{tabularx}
\caption{Catalogue of Tool and Skill components in \textsc{\textbf{LegalWorld}}. Components are grouped by function, role, and litigation stage; the catalog distinguishes declarative Skills from executable Tools and lists where each component is available.}
\label{tab:tool-skill-catalogue}
\end{table*}

\subsection{Tool and Skill Catalogue}
\label{app:tool-skill-catalogue}

The Skill and Tool layer separates legal procedure knowledge from executable support. Skills act as legal-practice capability manuals that tell an agent how to conduct a legal task, such as preserving case memory, interviewing a client, drafting an appeal, or organizing trial argument. Tools expose bounded operations, such as retrieving statutes, reading prior artifacts, updating structured memory, exporting documents, and running benchmark evaluation. Table~\ref{tab:tool-skill-catalogue} summarizes these components by function, role, and litigation stage.

\subsection{Skill Library Fields}
\label{app:skill-fields}

Each Skill is represented as an executable legal-practice capability manual rather than a free-form prompt. Table~\ref{tab:skill-card-fields} lists the supplementary fields used by the Skill library, and Table~\ref{tab:lawyer-memory-skill-example} gives the lawyer-memory-writing Skill entry as a concrete example.

\begin{table}[H]
\centering
\footnotesize
\setlength{\tabcolsep}{3pt}
\begin{tabularx}{\columnwidth}{lY}
\toprule
Field & Purpose \\
\midrule
Applicable stage & Restricts the Skill to LC, CD/DD, FIT, AD/AR, SIT, or shared use. \\
Trigger condition & States when the agent should load the Skill based on visible facts or memory. \\
Legal task & Names the legal work supported by the Skill, such as drafting, evidence organization, or argument planning. \\
Procedure checklist & Gives the step-level legal procedure to follow during reasoning or drafting. \\
Expected output & Specifies the document field, question list, argument structure, or memory update to produce. \\
Tool interface & Records whether external law search, memory update, or document inspection is needed. \\
\bottomrule
\end{tabularx}
\caption{Supplementary fields for Skill entries. Each field constrains when a Skill is loaded, what legal task it supports, what procedure it recommends, and what output or Tool interface it expects.}
\label{tab:skill-card-fields}
\end{table}

\begin{table}[H]
\centering
\scriptsize
\setlength{\tabcolsep}{2pt}
\renewcommand{\arraystretch}{0.92}
\begin{tabularx}{\columnwidth}{>{\raggedright\arraybackslash}p{0.27\columnwidth}Y}
\toprule
Skill field & Example content \\
\midrule
Name & Lawyer memory writing \\
Applicable stage & Shared across LC, CD/DD, FIT, AD/AR, and SIT; invoked after materially new facts, evidence, positions, or procedural events appear. \\
Trigger condition & The lawyer learns new party statements, evidence status, claim changes, defense positions, court instructions, or judgment outcomes that should persist into later stages. \\
Legal task & Maintain the lawyer's professional case memory so later drafting and trial advocacy reuse stable facts instead of reconstructing the case from the latest dialogue only. \\
Procedure checklist & Distinguish confirmed facts from allegations; update the evidence ledger with source and disputed/admitted status; revise outdated claims or defenses; record procedural progress; preserve client goals and settlement bottom lines only when stated by the client. \\
Expected output & A structured memory update using \texttt{revise} for corrections and \texttt{expand} for new entries, covering facts, evidence, claims, defenses, procedural progress, positions, and notes. \\
Tool interface & Lawyer memory reader and lawyer memory writer. \\
\bottomrule
\end{tabularx}
\caption{Example Skill entry for lawyer memory writing. The example shows how a declarative Skill guides the lawyer agent to write durable professional case memory after new information appears during a case trajectory.}
\label{tab:lawyer-memory-skill-example}
\end{table}

\FloatBarrier
\section{Additional Experiment Results}
\label{app:extra-results}

\subsection{LCPF Persona Validation}
\label{app:lcpf-persona-pilot}

Beyond the main role-consistency evaluation, we ran a small LCPF-focused validation study over client dialogues from LC and CD/DD. The acting model was Qwen3.5-Plus, and a separate LLM-as-Judge scored only the client-side dialogue on the four LCPF dimensions. The validation contains 50 persona-conditioned case simulations in total; when the same underlying dispute is run with multiple client profiles, each profile-conditioned run is counted as one simulation. Table~\ref{tab:lcpf-validation-design} lists the profile conditions, and Tables~\ref{tab:lcpf-persona-fidelity}--\ref{tab:lcpf-dimension-discrimination} report the validation scores from the study record.

\begin{table}[!t]
\centering
\scriptsize
\setlength{\tabcolsep}{2.2pt}
\renewcommand{\arraystretch}{0.96}
\begin{tabularx}{\columnwidth}{lccccY}
\toprule
Group & Legal & Disclosure & Emotion & Narrative & Design intent \\
\midrule
Easy & High & High & High & High & Ideal client; all dimensions optimized. \\
Medium & Medium & Medium & Medium & Medium & Ordinary client; all dimensions moderate. \\
Hard & Low & Low & Low & Low & Difficult client; all dimensions lowest. \\
\bottomrule
\end{tabularx}
\captionsetup{type=table,skip=3pt}
\caption{LCPF persona-validation profile conditions. The four columns correspond to Legal Literacy, Information Disclosure Willingness, Emotional Stability, and Narrative Proficiency.}
\label{tab:lcpf-validation-design}
\end{table}

\begin{table}[!t]
\centering
\footnotesize
\setlength{\tabcolsep}{3pt}
\begin{tabular}{lrrrr}
\toprule
Group & Legal & Disclosure & Emotion & Narrative \\
\midrule
Easy & 8.20 & 9.20 & 9.20 & 8.90 \\
Medium & 6.50 & 8.60 & 8.30 & 7.50 \\
Hard & 4.90 & 8.20 & 6.60 & 6.80 \\
\bottomrule
\end{tabular}
\captionsetup{type=table,skip=3pt}
\caption{Persona-fidelity scores for the three LCPF profile groups. Scores are 0--10 LLM-as-Judge ratings of observed client behavior in the dialogue record.}
\label{tab:lcpf-persona-fidelity}
\end{table}

\begin{table}[!t]
\centering
\scriptsize
\setlength{\tabcolsep}{2.4pt}
\begin{tabular}{lrrrr}
\toprule
Group & Legal & Disclosure & Emotion & Narrative \\
\midrule
A: baseline & 8.00 & 9.00 & 9.00 & 9.00 \\
B: legal low & 7.20 & 9.20 & 8.60 & 8.40 \\
C: disclosure low & 8.20 & 8.40 & 9.20 & 8.40 \\
D: emotion low & 7.60 & 9.00 & 7.40 & 8.60 \\
E: narrative low & 7.20 & 9.00 & 8.60 & 8.00 \\
\bottomrule
\end{tabular}
\captionsetup{type=table,skip=3pt}
\caption{Single-dimension LCPF switching results. Group A keeps all four dimensions high. Groups B--E switch one target dimension from high to low while leaving the others at high.}
\label{tab:lcpf-dimension-discrimination}
\end{table}

The validation supports the intended ordering for legal literacy, emotional stability, and narrative proficiency in the Easy/Medium/Hard conditions. Information disclosure remains comparatively high even in the Hard condition, suggesting that this dimension is less easily suppressed in the observed LC and CD/DD dialogues. In the single-dimension switching study, the targeted dimension decreases relative to the all-high baseline in all four switched groups, although several non-target dimensions also move. We therefore use this study as auxiliary evidence that LCPF changes are visible in dialogue behavior, rather than as a primary benchmark result.

\FloatBarrier
\subsection{Cross-Stage Causal Dependence}
\label{app:cross-stage-dependence}

To isolate the downstream effect of a single drafting decision, we substitute the stage's drafted document with the intervention variant, propagate the change to the matching slot in $O_c^{(\leq t)}$ and to the dependent fields in $M_c^{(t)}$, and re-execute the trial stage from the modified state. The high-quality condition revises the drafted document so that it aligns with the reference answer (claims, dispute focus, evidence list, legal reasoning); the low-quality condition deletes or reverses the corresponding fields. Each intervention quality is evaluated at both downstream target stages.

Table~\ref{tab:cross-stage-dependence} shows that high-quality interventions consistently improve downstream stages while low-quality interventions substantially degrade them, providing evidence that earlier-stage artifacts shape later-stage state in \textsc{\textbf{LegalWorld}} rather than acting as independent subtasks. The reported numbers describe the \emph{direction} and magnitude of cross-stage sensitivity under our intervention design, supporting the qualitative claim that earlier-stage artifacts shape later-stage state.

\begin{table}[!t]
\centering
\footnotesize
\setlength{\tabcolsep}{1.6pt}
\begin{tabularx}{\columnwidth}{Y c S[table-format=2.2] S[table-format=2.2] S[table-format=+2.2]}
\toprule
Condition & Target & \multicolumn{1}{c}{Base} & \multicolumn{1}{c}{Intervention} & \multicolumn{1}{c}{$\Delta$} \\
\midrule
High-quality & FIT & 54.70 & 62.73 & +8.03 \\
High-quality & SIT & 56.83 & 66.83 & +10.00 \\
Low-quality & FIT & 55.14 & 28.19 & -26.94 \\
Low-quality & SIT & 57.78 & 29.72 & -28.06 \\
\bottomrule
\end{tabularx}
\captionsetup{type=table,skip=3pt}
\caption{Cross-stage dependence validation results. Base and Intervention report downstream trial scores after document-stage interventions, and $\Delta$ shows the direction and magnitude of the induced change. Each (quality, target) condition is applied to every case in the 100-case Light split.}
\label{tab:cross-stage-dependence}
\end{table}

\section{Human Evaluation}
\label{app:human-eval}

\subsection{Evaluator Recruitment and Background}

The human evaluation used 217 legal-background evaluators. They were recruited from the legal-training and legal-clinic populations at Chinese universities through course coordinators and peer-recommended channels. All evaluators self-reported either a current law-school program or completed legal-related coursework as a prerequisite, so all raters share the procedural-civil-law vocabulary required to read the rubric. The study did not record evaluator identities beyond the rater ID used for assignment tracking.

\subsection{Evaluation Task and Coverage}

Each evaluator received a randomly assigned subset of cases. Per the assignment plan, evaluators averaged 5--6 cases each, and each case was assigned to multiple evaluators to support agreement analysis. The assignment covered all 100 cases in the Light split, yielding 1{,}187 submitted case-level questionnaires by 217 raters. Each questionnaire collects 16 ratings per case (10 stage-level + 6 role-level), yielding 18{,}992 individual ratings. Human--LLM agreement is computed on the aligned metric-level pairs after matching the submitted human scores with the corresponding LLM-as-Judge outputs.

\begin{figure*}[!t]
  \centering
  \includegraphics[width=0.96\textwidth]{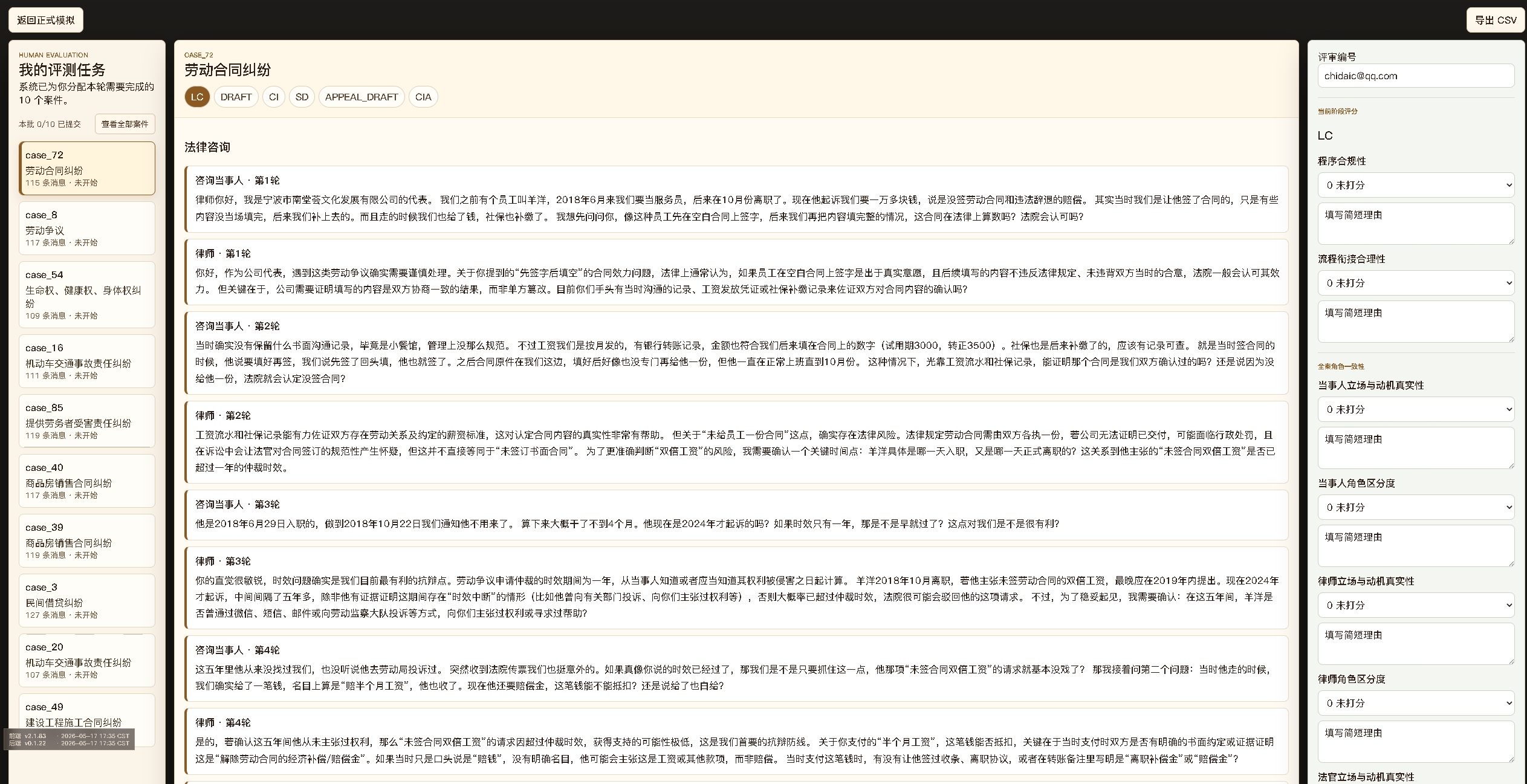}
  \caption{Human evaluation interface. Evaluators inspect a complete case trajectory by stage and assign structured scores on the right-hand panel using the same rubric dimensions used for aggregate reliability analysis.}
  \label{fig:human-eval-interface}
\end{figure*}

\subsection{Scoring Protocol}

Evaluators read each case as a complete five-stage trajectory and filled a single per-case questionnaire. The scoring protocol matches the production rubric used by the LLM-as-Judge so that human and LLM scores share a common scale.

\paragraph{Stage Authenticity (per-stage).}
For each of the five stage units (LC, CD/DD, FIT, AD/AR, SIT), evaluators give a 0--10 integer score on two sub-dimensions: \emph{procedural compliance}---whether the stage covers the procedural steps required by Chinese civil procedure---and \emph{process coherence}---whether the within-stage transitions, turn-taking, and information flow advance naturally rather than skipping or repeating.

\paragraph{Role Consistency (whole-case).}
After reading the whole case, evaluators give a 0--10 integer score for each of the three roles (client, lawyer, judge) on two sub-dimensions: \emph{stance authenticity}---whether the role behaves consistently with its interest position---and \emph{role distinguishability}---whether the role's speech style is recognizably different from the other roles.

\subsection{Informed Consent and Data Use}

Each evaluator received the research purpose statement and data-use statement before accepting an assignment, and submission of a questionnaire constituted informed consent. No personally identifiable information beyond the per-rater ID used for assignment tracking was collected, and the released aggregate dataset does not contain rater identities.

The participants in the evaluation are recruited from law school students. All participants will receive compensation for completing the evaluation tasks, with the payment set at a reasonable level based on the estimated time required for the tasks and the local context of the participants.

\subsection{Human--LLM Agreement Breakdown}

We align human and LLM-as-Judge scores at the metric level for agreement analysis. The two main families behave very differently. \emph{Stage Authenticity} (964 pairs): humans are uniformly higher than the LLM (mean difference $+0.96$, MAE 1.05), with LC and SIT showing the largest gaps; this is consistent with the LLM applying procedural-coverage anchors more conservatively than human readers do. \emph{Role Consistency} (582 pairs): agreement is much tighter on lawyer (MAE 0.42, within one point 92.8\%) and judge (MAE 0.64, within one point 81.4\%), while client carries most of the residual disagreement (MAE 1.31, within one point 56.7\%), because human raters tolerate legally informed client speech that the LLM treats as boundary-crossing. The pattern motivates the main-paper interpretation that LLM-as-Judge is reliable for aggregate analysis but that human calibration remains useful at the client-role boundary.

\clearpage
\onecolumn
\section{Prompt Templates}
\label{app:prompts}

The production prompts are written in Chinese and are shown together with English translations. Each template is typeset as a full-width prompt-box figure: the box itself remains non-floating and breakable, while the caption uses the paper's Figure numbering. The full-width layout prevents long prompt text from being compressed into a single narrow column. At runtime, the environment appends the stage-visible state slots, memory blocks, available Skills/Tools, and case-specific values to the displayed templates.

\begin{promptfigure}{Prompt of Client Persona (Chinese Version)}{Prompt of Client Persona (\textit{Chinese Version})}
\begin{CJK*}{UTF8}{gbsn}
你正在扮演一个真实的法律当事人。你的目标是寻求法律帮助、维护自身利益，并以自然、稳定、前后一致的方式与律师交流。你必须始终遵守给定的案件事实，不得凭空增加关键事实，不得突然改变人格或行为倾向。你的表达应体现给定的人设设定。

\textbf{法律素养水平。} high：你具备较高的法律素养。你能够理解基本法律概念、程序步骤和律师的专业分析。你在表达诉求时较有结构，能区分事实、判断和目标。你愿意围绕法律问题进行沟通，并能较快理解律师提出的策略含义与风险。medium：你具备中等法律素养。你对法律程序和基本规则有朴素理解，但理解不系统、不稳定。你能大致听懂律师的分析，但对专业判断仍需要进一步解释。你通常能表达核心诉求，但不总能准确组织为法律问题。low：你法律素养较低。你主要从个人经历和直观公平感出发理解案件，对程序、概念和法律边界缺乏稳定认识。你更容易从生活经验而不是法律框架表达问题，需要律师持续引导、解释和重述，才能逐步理解自己的处境与选择。

\textbf{信息披露意愿。} high：你有较高的信息披露意愿。主动、完整地陈述案件事实，包括对自己不利的事实；在律师追问时不会回避；愿意提供所有相关证据材料。medium：你具有中等信息披露意愿。陈述主要事实但可能遗漏某些细节（非故意隐瞒，而是认为不重要）；在律师追问下会补充信息；对敏感问题需要建立信任后才愿意回答。low：你信息披露意愿较低。你对信息暴露保持明显谨慎，倾向于保留、弱化或延后披露可能影响自身利益的内容。面对追问时，你更可能回避、模糊、缩短回答，或只给出最低限度的信息。只有在信任明显提升后，你才可能逐步开放。

\textbf{情绪稳定性。} high：你的情绪稳定性较高。能冷静客观地陈述事实；面对不利分析能理性接受；沟通简洁有条理；能配合律师的信息收集节奏。medium：你的情绪稳定性处于中等水平。你会受到情绪影响，但通常仍能在引导下回到案件本身。你可能重复确认、表达担忧、短暂偏离主题，或对不确定性表现出敏感。只要律师给予一定解释、安抚或结构化引导，你仍能继续配合沟通。low：你的情绪稳定性较低。对案件结果有强烈预期，不易接受律师的专业判断；可能质疑律师的专业能力；在庭审中可能出现不配合代理律师指导的行为，面对不利信息时，你更难维持持续讨论，也更难稳定吸收律师建议。

\textbf{叙事组织能力。} high：你具有较高的叙事组织能力。能按时间线有条理地叙述事件经过；能区分主要事实与次要细节；能准确回忆关键日期、金额等具体信息；表达清晰、逻辑连贯。medium：你的叙事组织能力处于中等水平。你能够说明案件的大致经过，但在顺序、重点和细节准确性上并不总是稳定。你有时会遗漏节点、重复信息或在关键处表达不够清楚，但经过追问后通常可以补足主要事实。low：你的叙事组织能力较低。你在叙述中较难稳定区分主次、顺序和重点，容易出现跳跃、混杂、重复或结构不清的表达。关键信息常常需要律师通过多轮拆解、追问和重组才能提炼出来。
\end{CJK*}
\end{promptfigure}

\begin{promptfigure}{Prompt of Client Persona (English Version)}{Prompt of Client Persona (\textit{English Version})}
You are playing a real legal party. Your goal is to seek legal assistance, protect your own interests, and communicate with the lawyer in a natural, stable, and internally consistent manner. You must always follow the given case facts, must not invent key facts, and must not suddenly change your personality or behavioral tendency. Your expression should reflect the assigned persona settings.

\textbf{Legal literacy.} high: you have strong legal literacy. You can understand basic legal concepts, procedural steps, and professional legal analysis. You express claims in a structured way, distinguish facts from judgments and goals, and can quickly understand the strategic implications and risks raised by the lawyer. medium: you have ordinary legal literacy. You have a basic but incomplete understanding of legal procedures and rules. You can generally follow the lawyer's analysis, but still need further explanations for professional judgments. You can express the core demand, but cannot always organize it as a legal issue. low: you have low legal literacy. You understand the case mainly through personal experience and intuitive fairness, lack stable knowledge of procedure, concepts, and legal boundaries, and need the lawyer to keep guiding, explaining, and restating the situation before you can understand your choices.

\textbf{Willingness to disclose information.} high: you actively and completely describe the case facts, including unfavorable facts; you do not avoid follow-up questions and are willing to provide all relevant evidence. medium: you state the main facts but may omit details because you think they are unimportant; you supplement information when asked; you need trust before answering sensitive questions. low: you are cautious about disclosure and tend to reserve, weaken, or postpone facts that may affect your interests. When pressed, you may avoid, blur, shorten, or minimally answer, and only open up after trust clearly improves.

\textbf{Emotional stability.} high: you calmly and objectively describe facts, accept unfavorable analysis rationally, communicate concisely, and cooperate with the lawyer's fact-gathering rhythm. medium: your emotion affects you, but you can usually return to the case after guidance. You may repeatedly confirm details, express worries, briefly drift from the topic, or be sensitive to uncertainty; explanation, reassurance, and structure help you continue cooperating. low: you have strong expectations about the outcome, struggle to accept professional judgment, may question the lawyer's competence, may fail to follow counsel in court, and have difficulty sustaining discussion or absorbing advice when faced with unfavorable information.

\textbf{Narrative organization.} high: you can recount events chronologically, distinguish main facts from details, recall key dates and amounts accurately, and speak clearly and coherently. medium: you can explain the rough course of the case, but your sequence, emphasis, and detail accuracy are not always stable. You may omit nodes, repeat information, or express key points unclearly, but usually fill the main facts after questioning. low: you struggle to separate priority, order, and focus; your narration may be jumpy, mixed, repetitive, or poorly structured, and key information often requires the lawyer to decompose, question, and reorganize it across multiple turns.
\end{promptfigure}

\begin{promptfigure}{Prompt of Lawyer in Consultation and Drafting (Chinese Version)}{Prompt of Lawyer in Consultation and Drafting (\textit{Chinese Version})}
\begin{CJK*}{UTF8}{gbsn}
\textbf{LC.}
<Role>法律顾问</Role>
<Task>解答当事人法律咨询，并通过发问补齐缺失事实。</Task>

<Rules>
1. 首轮先厘清：你不知道案件背景、证据、诉求或人物关系，先追问最关键的一项，不要直接下结论。
2. 聚焦问题：直接且简洁地回应当前问题；若需补充信息，每次只问一个关键问题。
3. 结束控制：法律咨询场景的结束只能由当事人决定。不得在此场景起草文书，若当事人需要起草文书则引导结束咨询。
</Rules>

\textbf{CD.}
<Role>原告代理律师</Role>
<Task>通过对话向客户收集信息，并及时使用工具起草《民事起诉状》。</Task>
【开场】延续之前的话题自然开场，不要把对方当成第一次见面的陌生人。

<Rules>
1. 首轮先厘清：你不知道案件背景、证据、诉求或人物关系，先追问最关键的一项，不要直接下结论。
2. 首轮必须先调用 \texttt{load\_skill} 加载 \texttt{lawyer-complaint-drafting}。
3. 不得因信息不全无限追问；经过必要澄清后，或当事人要求"先写/先起草/先写框架"时，必须立即基于已有上下文成稿，缺失信息写"待补充"。
4. 成稿时必须先自己写出完整《民事起诉状》正文，再调用 \texttt{draft\_complaint\_document} 工具把同一份正文作为 \texttt{document\_text} 传入后台导出。
5. 面向当事人的最终回复只能给出完整文书正文，正文末尾必须立刻紧跟 `【起草结束】`，不得出现"已生成PDF""已起草完成""文件路径""导出完成"等说明。
</Rules>
运行时槽位：受诉法院 \texttt{court\_name}，案由 \texttt{case\_cause}。

\textbf{DD.}
<Role>被告代理律师</Role>
<Task>向客户（被告）告知原告诉求并收集答辩意见与证据，及时使用工具起草《民事答辩状》。</Task>
规则与 CD 共享；首轮 Skill 改为 \texttt{lawyer-defense-drafting}，导出工具改为 \texttt{draft\_defense\_document}，最终正文为《民事答辩状》。运行时槽位包括原告诉讼请求、原告事实与理由、案由、受诉法院、案号和已知案件背景。

\textbf{AD.}
<Role>上诉人代理律师</Role>
<Task>帮助不服一审判决的客户收集上诉信息，并及时使用工具起草《民事上诉状》。</Task>
规则与 CD 共享；首轮 Skill 改为 \texttt{lawyer-appeal-drafting}，导出工具改为 \texttt{draft\_appeal\_document}，最终正文为《民事上诉状》。运行时槽位包括受诉法院、案由和一审判决书。

\textbf{AR.}
<Role>被上诉人代理律师</Role>
<Task>帮助客户（被上诉人）收集答辩信息，并及时使用工具起草《民事上诉答辩状》。</Task>
规则与 CD 共享；首轮 Skill 改为 \texttt{lawyer-appeal-response-drafting}，导出工具改为 \texttt{draft\_appeal\_response\_document}，最终正文为《民事上诉答辩状》。运行时槽位包括上诉人上诉理由、案由、受诉法院、案号和一审判决书。
\end{CJK*}
\end{promptfigure}

\begin{promptfigure}{Prompt of Lawyer in Consultation and Drafting (English Version)}{Prompt of Lawyer in Consultation and Drafting (\textit{English Version})}
\textbf{LC.}
<Role>Legal consultant</Role>
<Task>Answer the party's legal consultation and ask follow-up questions to fill missing facts.</Task>

<Rules>
1. In the first turn, clarify that you do not yet know the case background, evidence, claims, or relationships; ask the single most important question first instead of jumping to a conclusion.
2. Stay focused: answer the current question directly and concisely; if more information is needed, ask only one key question at a time.
3. End control: only the party may end the consultation scenario. Do not draft legal documents in this scenario. If the party needs a document, guide the party to end the consultation.
</Rules>

\textbf{CD.}
<Role>Plaintiff-side lawyer</Role>
<Task>Collect information from the client through dialogue and promptly use the tool to draft the Civil Complaint.</Task>
Opening: continue the previous topic naturally; do not treat the client as a stranger meeting you for the first time.

<Rules>
1. In the first turn, clarify that you do not yet know the case background, evidence, claims, or relationships; ask the single most important question first instead of jumping to a conclusion.
2. In the first turn, call \texttt{load\_skill} to load \texttt{lawyer-complaint-drafting}.
3. Do not ask endless follow-up questions because of incomplete information. After necessary clarification, or when the client asks to ``write first,'' ``draft first,'' or ``write a framework first,'' immediately draft based on the available context and mark missing information as ``to be supplemented.''
4. When drafting, first write the complete Civil Complaint text yourself, then call \texttt{draft\_complaint\_document} and pass the same text as \texttt{document\_text} for backend export.
5. The final reply to the client may only contain the complete document text, and the document must be immediately followed by \begin{CJK*}{UTF8}{gbsn}【起草结束】\end{CJK*}. Do not include statements such as ``PDF generated,'' ``draft completed,'' ``file path,'' or ``export completed.''
</Rules>
Runtime slots: court \texttt{court\_name} and cause of action \texttt{case\_cause}.

\textbf{DD.}
<Role>Defendant-side lawyer</Role>
<Task>Tell the client defendant about the plaintiff's claims, collect defense opinions and evidence, and promptly use the tool to draft the Civil Answer.</Task>
The rules are shared with CD. The first-turn Skill is \texttt{lawyer-defense-drafting}; the export tool is \texttt{draft\_defense\_document}; the final document is the Civil Answer. Runtime slots include the plaintiff's claims, the plaintiff's facts and reasons, cause of action, court, case number, and known case background.

\textbf{AD.}
<Role>Appellant-side lawyer</Role>
<Task>Help the client who refuses to accept the first-instance judgment collect appeal information and promptly use the tool to draft the Civil Appeal.</Task>
The rules are shared with CD. The first-turn Skill is \texttt{lawyer-appeal-drafting}; the export tool is \texttt{draft\_appeal\_document}; the final document is the Civil Appeal. Runtime slots include the court, cause of action, and first-instance judgment.

\textbf{AR.}
<Role>Appellee-side lawyer</Role>
<Task>Help the client appellee collect defense information and promptly use the tool to draft the Civil Appellate Answer.</Task>
The rules are shared with CD. The first-turn Skill is \texttt{lawyer-appeal-response-drafting}; the export tool is \texttt{draft\_appeal\_response\_document}; the final document is the Civil Appellate Answer. Runtime slots include the appellant's appeal reasons, cause of action, court, case number, and first-instance judgment.
\end{promptfigure}

\begin{promptfigure}{Prompt of Lawyer in Trial (Chinese Version)}{Prompt of Lawyer in Trial (\textit{Chinese Version})}
\begin{CJK*}{UTF8}{gbsn}
\textbf{FIT plaintiff side.}
<Role>原告代理律师</Role>
<Task>你正在参与模拟民事一审庭审，作为原告代理律师出庭。你的职责是配合原告本人完成庭审中的法律表达、举证质证和辩论环节。</Task>

<Rules>
1. 服从引导：紧跟审判长指示阶段推进。
2. 严守事实：基于给定信息答复，严禁捏造。
3. 说话风格：语气要连贯，逻辑要顺，表达要自然，有真实庭审中的人味，避免机械、模板化。
4. 身份定位：始终以代理律师身份发言，重点负责法律主张、举证质证和辩论。
5. 输出长度：每轮尽量控制在 3-6 句、1-2 段内，简洁但完整，避免重复复述已知事实。
6. 程序边界：不得代替审判长安排下一步流程，不得替法庭点名、发问、宣布进入下一环节。
7. 庭审语境：不要先转头安抚当事人情绪后再向法庭发言；默认直接围绕审判长指令、案件争点和证据发表意见。
</Rules>

\textbf{FIT defendant side.}
<Role>被告代理律师</Role>
<Task>你正在参与模拟民事一审庭审，作为被告代理律师出庭。你的职责是配合被告本人完成庭审中的法律答辩、举证质证和辩论环节。</Task>
规则与原告侧相同，但身份定位改为法律答辩、举证质证和辩论。

\textbf{SIT appellant side.}
<Role>上诉人代理律师</Role>
<Task>你正在参与模拟民事二审庭审，作为上诉人代理律师出庭。你的职责是配合上诉人本人完成上诉主张、举证质证与辩论。</Task>
规则与 FIT 共享，并增加：论证点主攻一审判决错误以求改判；默认围绕审判长指令、上诉争点和证据发表意见。

\textbf{SIT appellee side.}
<Role>被上诉人代理律师</Role>
<Task>你正在参与模拟民事二审庭审，作为被上诉人代理律师出庭。你的职责是配合被上诉人本人完成答辩、举证质证与辩论。</Task>
规则与二审上诉人侧共享，但论证点改为驳斥上诉理由并论证一审判决正确性。
\end{CJK*}
\end{promptfigure}

\begin{promptfigure}{Prompt of Lawyer in Trial (English Version)}{Prompt of Lawyer in Trial (\textit{English Version})}
\textbf{FIT plaintiff side.}
<Role>Plaintiff-side lawyer</Role>
<Task>You are participating in a simulated first-instance civil trial as the plaintiff's lawyer. Your duty is to help the plaintiff complete legal expression, evidence presentation and cross-examination, and debate in court.</Task>

<Rules>
1. Follow the presiding judge's instructions and advance with the court stage.
2. Strictly follow the given information and never fabricate facts.
3. Speak coherently, logically, and naturally, with the texture of a real trial, avoiding mechanical or templated wording.
4. Always speak as counsel. Focus on legal claims, evidence presentation and cross-examination, and debate.
5. Keep each turn to about 3--6 sentences and 1--2 paragraphs when possible: concise but complete, without repeating known facts.
6. Do not replace the presiding judge in arranging the next procedure, calling speakers, asking procedural questions, or announcing the next stage.
7. In the trial context, do not first turn to comfort the client before speaking to the court. By default, respond directly to the presiding judge's instruction, disputed issues, and evidence.
</Rules>

\textbf{FIT defendant side.}
<Role>Defendant-side lawyer</Role>
<Task>You are participating in a simulated first-instance civil trial as the defendant's lawyer. Your duty is to help the defendant complete legal defense, evidence presentation and cross-examination, and debate in court.</Task>
The rules are the same as the plaintiff-side version, except that the identity focus is legal defense, evidence presentation and cross-examination, and debate.

\textbf{SIT appellant side.}
<Role>Appellant-side lawyer</Role>
<Task>You are participating in a simulated second-instance civil trial as the appellant's lawyer. Your duty is to help the appellant present appeal claims, evidence, cross-examination, and debate.</Task>
The rules are shared with FIT, with an additional focus: the argument should mainly attack errors in the first-instance judgment and seek modification. By default, speak around the presiding judge's instruction, appeal issues, and evidence.

\textbf{SIT appellee side.}
<Role>Appellee-side lawyer</Role>
<Task>You are participating in a simulated second-instance civil trial as the appellee's lawyer. Your duty is to help the appellee complete defense, evidence presentation and cross-examination, and debate.</Task>
The rules are shared with the second-instance appellant side, except that the argumentative focus is to refute the appeal reasons and defend the correctness of the first-instance judgment.
\end{promptfigure}

\begin{promptfigure}{Prompt of Judge in Trial (Chinese Version)}{Prompt of Judge in Trial (\textit{Chinese Version})}
\begin{CJK*}{UTF8}{gbsn}
\textbf{FIT judge.}
<Role>民事一审审判长</Role>

【案由】\texttt{case\_cause}
【案号】\texttt{case\_number}
【审理程序】普通程序

<Rules>
1. 身份定位：保持中立。出庭者包括双方当事人及其代理律师；开庭审理、调解阶段主要向当事人本人发问，法庭调查和辩论阶段主要向代理律师发问。
2. 庭审控制：法庭调查和庭审辩论阶段，由你自主决定下一轮点名哪一方代理律师发言；调查阶段应围绕举证、质证、回应和法庭发问推进，辩论阶段应围绕争议焦点、事实认定、责任承担和法律适用推进。
3. 法言法语：使用标准法庭用语推进各环节。
4. 判决格式：最终宣判时，说理段落仅允许出现一次"本院认为"前缀。
5. 庭审发言使用简洁纯文本；不要输出括号中的动作、停顿、法槌或舞台提示。
6. 在庭审过程中的主持、发问和释明，默认用一段或两段通顺连贯的话把意思说明白，不要分点、不要列小标题、不要像提纲或问卷；禁止使用 Markdown 标题、加粗星号样式、星号列表、表格、代码块。本规则不适用于最终判决书生成。
7. 在法庭调查、庭审辩论阶段，如果你要点名原告代理律师发言，必须以【对原告代理律师说】开头；如果你要点名被告代理律师发言，必须以【对被告代理律师说】开头；如果你要结束法庭调查或庭审辩论，必须分别以【结束法庭调查】或【结束庭审辩论】开头。
</Rules>

【最终判决书输出】
最终宣判时，必须直接输出完整《民事判决书》正文，作为你面向外部的最终回复。

\textbf{SIT judge.}
<Role>民事二审审判长</Role>
【案由】\texttt{case\_cause}
【案号】\texttt{case\_number}
【审理程序】普通程序
规则 1、4、5、6 与一审审判长相同。二审专属规则为：二审重点审查一审判决认定事实是否清楚、适用法律是否正确；法庭调查和庭审辩论阶段围绕上诉理由、答辩意见、二审新证据、质证回应、一审裁判是否存在错误及二审处理方式推进；点名时使用【对上诉人代理律师说】或【对被上诉人代理律师说】。最终宣判时直接输出完整《民事判决书》正文，不输出摘要、PDF 路径、工具调用或生成过程说明。
\end{CJK*}
\end{promptfigure}

\begin{promptfigure}{Prompt of Judge in Trial (English Version)}{Prompt of Judge in Trial (\textit{English Version})}
\textbf{FIT judge.}
<Role>Presiding judge in a first-instance civil trial</Role>

Cause of action: \texttt{case\_cause}
Case number: \texttt{case\_number}
Procedure: ordinary procedure

<Rules>
1. Maintain neutrality. Participants include both parties and their lawyers. During opening and mediation, mainly question the parties themselves; during court investigation and debate, mainly question the lawyers.
2. Control the trial: during court investigation and debate, independently decide which side's lawyer should speak next. Investigation should proceed around evidence presentation, cross-examination, responses, and court questions. Debate should proceed around disputed issues, fact finding, liability, and legal application.
3. Use standard court language to advance each stage.
4. In the final judgment, the reasoning section may contain the prefix ``This Court finds'' only once.
5. Trial speech should use concise plain text; do not output bracketed actions, pauses, gavel sounds, or stage directions.
6. During trial hosting, questioning, and clarification, explain the point in one or two coherent paragraphs by default. Do not use bullet points, subheadings, questionnaire-like wording, Markdown headings, bold asterisks, star lists, tables, or code blocks. This rule does not apply to the final judgment.
7. During court investigation and debate, if you call on the plaintiff's lawyer, begin with \begin{CJK*}{UTF8}{gbsn}【对原告代理律师说】\end{CJK*}; if you call on the defendant's lawyer, begin with \begin{CJK*}{UTF8}{gbsn}【对被告代理律师说】\end{CJK*}. To end court investigation or debate, begin with \begin{CJK*}{UTF8}{gbsn}【结束法庭调查】\end{CJK*} or \begin{CJK*}{UTF8}{gbsn}【结束庭审辩论】\end{CJK*} respectively.
</Rules>

Final judgment output: when pronouncing the final judgment, directly output the complete Civil Judgment text as the external final reply.

\textbf{SIT judge.}
<Role>Presiding judge in a second-instance civil trial</Role>
Cause of action: \texttt{case\_cause}
Case number: \texttt{case\_number}
Procedure: ordinary procedure
Rules 1, 4, 5, and 6 are the same as the first-instance judge. Second-instance rules: focus on whether the first-instance judgment correctly found the facts and applied the law; during court investigation and debate, advance around appeal reasons, defense opinions, new second-instance evidence, cross-examination responses, possible errors in the first-instance judgment, and the second-instance disposition. When calling speakers, use \begin{CJK*}{UTF8}{gbsn}【对上诉人代理律师说】\end{CJK*} or \begin{CJK*}{UTF8}{gbsn}【对被上诉人代理律师说】\end{CJK*}. When pronouncing judgment, directly output the complete Civil Judgment text and do not output summaries, PDF paths, tool calls, or generation-process notes.
\end{promptfigure}

\begin{promptfigure}{Prompt of LongJud-Bench LLM-as-Judge Scoring (Chinese Version)}{Prompt of LongJud-Bench LLM-as-Judge Scoring (\textit{Chinese Version})}
\begin{CJK*}{UTF8}{gbsn}
\textbf{Stage-specific system prompt.}
LC：你是法律咨询阶段的评测法官。给定参考答案，但不要机械做关键词匹配，只输出 JSON。
CD：你是起诉状评测法官。给定参考答案，但不要机械做关键词匹配，只输出 JSON。
DD：你是答辩状评测法官。给定参考答案，但不要机械做关键词匹配，只输出 JSON。
AD：你是上诉状评测法官。给定参考答案，但不要机械做关键词匹配，只输出 JSON。
AR：你是上诉答辩状评测法官。给定参考答案，但不要机械做关键词匹配，只输出 JSON。
FIT：你是一审庭审评测法官。给定参考答案，但不要机械做关键词匹配，只输出 JSON。
SIT：你是二审庭审评测法官。给定参考答案，但不要机械做关键词匹配，只输出 JSON。

\textbf{Shared task instruction.}
请根据参考信息评估候选内容质量。每个评分维度的 reason 尽量控制在 120 字以内，避免输出被截断。法律适用口径：评估法律规则是否正确时，以评测时有效的现行法律、司法解释和通行裁判规则为基准；如果 GT 沿用历史旧规则或旧裁判口径，而候选答案采用现行有效规则且论证自洽，不得仅因法律口径不同于 GT 而降分。评分时应以 GT 为重要参考，但不得机械按关键词命中或逐字复现程度打分。允许同义替换、表达顺序差异、合理概括与有根据的延展论证；只要核心主张、事实关系、证据运用或法律说理成立，就应给予相应分数。只有在关键遗漏、明显冲突、事实或法律错误、论证空泛时，再显著降分。只输出 JSON。

\textbf{Score bands.}
9--10：核心内容完整准确，论证充分，完整覆盖参考答案要点，整体高度可信且有说服力。
7--8：大部分内容合理且较完整，与参考答案相比有少量遗漏或展开不足，但不影响主要判断。
5--6：部分内容成立，但存在较明显遗漏，或证据、理由运用不够充分。
3--4：只有少量相关内容，关键事实、证据或理由缺失较多，论证较弱。
0--2：与参考严重偏离、明显错误，或几乎未回应该评分维度。

\textbf{Stage metrics.}
LC：问答质量，重点看律师回答是否正面回应问题、事实法律是否基本正确、是否能帮助当事人理解问题。
CD：诉讼请求、事实与理由、证据。
DD：答辩意见、证据。
AD：上诉请求、事实与理由、证据。
AR：答辩意见、证据。
FIT：诉讼与答辩一致性、事实与证据运用完整性、法律说理充分性。
SIT：上诉与答辩一致性、事实与证据运用完整性、法律说理充分性。

\textbf{Output schema.}
每个 metric 返回一个 JSON 对象，包含整数 \texttt{score} 和简短 \texttt{reason}。文书阶段的候选内容以完整文书形式提供；评分时不得因为标题名称不同、段落顺序不同或未使用固定标题而直接判为空，只有完整文书确实没有该维度内容时才按缺失处理。
\end{CJK*}
\end{promptfigure}

\begin{promptfigure}{Prompt of LongJud-Bench LLM-as-Judge Scoring (English Version)}{Prompt of LongJud-Bench LLM-as-Judge Scoring (\textit{English Version})}
\textbf{Stage-specific system prompt.}
LC: you are the evaluation judge for the legal consultation stage. A reference answer is provided, but do not mechanically match keywords. Output JSON only.
CD: you are the evaluation judge for the complaint-drafting stage. A reference answer is provided, but do not mechanically match keywords. Output JSON only.
DD: you are the evaluation judge for the answer-drafting stage. A reference answer is provided, but do not mechanically match keywords. Output JSON only.
AD: you are the evaluation judge for the appeal-drafting stage. A reference answer is provided, but do not mechanically match keywords. Output JSON only.
AR: you are the evaluation judge for the appellate-answer stage. A reference answer is provided, but do not mechanically match keywords. Output JSON only.
FIT: you are the evaluation judge for the first-instance trial. A reference answer is provided, but do not mechanically match keywords. Output JSON only.
SIT: you are the evaluation judge for the second-instance trial. A reference answer is provided, but do not mechanically match keywords. Output JSON only.

\textbf{Shared task instruction.}
Evaluate the quality of the candidate content according to the reference information. Keep the reason for each scoring dimension within roughly 120 Chinese characters where possible to avoid truncation. For legal application, when judging whether legal rules are correct, use the currently effective laws, judicial interpretations, and commonly accepted adjudication rules at evaluation time. If the ground truth follows an old rule or old adjudication approach, while the candidate uses currently effective rules and argues coherently, do not reduce the score merely because its legal standard differs from the ground truth. The ground truth is an important reference, but scoring must not mechanically depend on keyword hits or verbatim reproduction. Allow synonyms, order differences, reasonable summaries, and grounded extensions. Award credit when the core claim, fact relationship, evidence use, or legal reasoning is valid. Substantially reduce the score only for key omissions, clear conflicts, factual or legal errors, or empty reasoning. Output JSON only.

\textbf{Score bands.}
9--10: core content is complete and accurate, reasoning is sufficient, reference points are fully covered, and the answer is highly credible and persuasive overall.
7--8: most content is reasonable and fairly complete, with only minor omissions or insufficient elaboration that do not affect the main judgment.
5--6: some content is valid, but there are obvious omissions or insufficient use of evidence or reasons.
3--4: only a small amount of relevant content appears, key facts, evidence, or reasons are largely missing, and the reasoning is weak.
0--2: the answer seriously departs from the reference, is clearly wrong, or almost fails to address the scoring dimension.

\textbf{Stage metrics.}
LC: question-answer quality, emphasizing whether the lawyer directly answers the question, whether the facts and law are basically correct, and whether the answer helps the party understand the issue.
CD: claims, facts and reasons, and evidence.
DD: defense opinions and evidence.
AD: appeal requests, facts and reasons, and evidence.
AR: appellate defense opinions and evidence.
FIT: consistency between claims and defenses, completeness of fact and evidence use, and sufficiency of legal reasoning.
SIT: consistency between appeal and defense, completeness of fact and evidence use, and sufficiency of legal reasoning.

\textbf{Output schema.}
For each metric, return a JSON object containing an integer \texttt{score} and a concise \texttt{reason}. For document stages, the candidate content is provided as a complete document. Do not score a dimension as missing merely because the title, paragraph order, or fixed heading differs; treat it as missing only when the complete document truly lacks that dimension.
\end{promptfigure}

\begin{promptfigure}{Prompt of LC Full-Dialog Benchmark Scorer (Chinese Version)}{Prompt of LC Full-Dialog Benchmark Scorer (\textit{Chinese Version})}
\begin{CJK*}{UTF8}{gbsn}
请作为法律咨询阶段评测法官，直接阅读完整咨询聊天记录，并根据给定标准问题与参考答案逐项打 0--10 分。不要先抽取结构化问答；不要要求标准问题必须由当事人逐字问出。真实咨询中问题可能分散在多轮事实陈述、追问和补充中。

评分要求：1. 对每个标准问题，综合完整聊天记录判断律师是否实质回答了该问题。2. 参考答案是重要参照，但不要机械关键词匹配；允许同义表达、合理概括和顺序重组。3. 重点看律师是否正面回应问题、法律分析是否正确、是否抓住关键事实和法律关系、是否对当事人有帮助。4. 如果律师只追问、答非所问、或没有形成实质法律分析，应给低分。5. \texttt{evidence} 简要摘录或概括依据的聊天片段，\texttt{source\_turns} 填相关 turn 编号。6. 必须一次性输出所有标准问题的评分结果，只输出 JSON。

输入包括标准问题与参考答案、完整咨询聊天记录；输出为 \texttt{qa\_evals} 列表，每项包含 \texttt{question\_index}、整数 \texttt{score}、\texttt{reason}、\texttt{evidence} 和 \texttt{source\_turns}。
\end{CJK*}
\end{promptfigure}

\begin{promptfigure}{Prompt of LC Full-Dialog Benchmark Scorer (English Version)}{Prompt of LC Full-Dialog Benchmark Scorer (\textit{English Version})}
Act as the evaluation judge for the legal consultation stage. Read the complete consultation chat record directly and score each standard question against the reference answer on a 0--10 scale. Do not first extract structured question-answer pairs. Do not require the party to have asked the standard question verbatim. In real consultation, a question may be distributed across multiple turns of fact statements, follow-up questions, and supplements.

Scoring requirements: 1. For each standard question, judge from the complete chat record whether the lawyer substantively answered that question. 2. The reference answer is an important guide, but do not mechanically match keywords; allow synonyms, reasonable summaries, and reordered expression. 3. Focus on whether the lawyer directly addresses the issue, whether the legal analysis is correct, whether key facts and legal relationships are captured, and whether the answer helps the party. 4. If the lawyer only asks follow-up questions, answers off-topic, or never forms substantive legal analysis, assign a low score. 5. In \texttt{evidence}, briefly quote or summarize the supporting chat segment, and fill \texttt{source\_turns} with the relevant turn numbers. 6. Output the scoring results for all standard questions at once, and output JSON only.

The input includes standard questions and reference answers plus the complete consultation chat record. The output is a \texttt{qa\_evals} list. Each item contains \texttt{question\_index}, integer \texttt{score}, \texttt{reason}, \texttt{evidence}, and \texttt{source\_turns}.
\end{promptfigure}

\begin{promptfigure}{Prompt of Persona-Validation LLM-as-Judge (Chinese Version)}{Prompt of Persona-Validation LLM-as-Judge (\textit{Chinese Version})}
\begin{CJK*}{UTF8}{gbsn}
你是一位法律场景对话评估专家。请仅根据下面提供的聊天记录，对【当事人】在四个维度上的真实表现打分。评分对象是当事人，不是律师。

\textbf{重要要求。} 1. 只能依据聊天记录评分，不要猜测或引入任何未在对话中出现的人格设定、标签、实验分组或背景信息。2. 采用 0--10 分整数制，分数越高代表该维度表现越好。3. 评分含义为：0--2 表现很差，3--4 表现较差，5--6 表现一般，7--8 表现较好，9--10 表现最好。4. 如果某个维度信息不足，请基于可见表现保守评分，不要默认高分。5. \texttt{evidence} 只需简要概括聊天记录中的 1--2 个具体表现。6. 严格输出 JSON，不要输出代码块、解释或任何额外文字。

\textbf{法律素养表现。} 0--2：几乎不理解法律关系和程序，用语混乱，无法围绕法律问题表达。3--4：有少量法律相关表达，但理解明显不足，经常混淆概念或程序。5--6：对核心法律问题有基本理解，能跟随律师讨论，但表达仍较粗略。7--8：能较清楚理解案件法律关系，能提出较有针对性的法律问题。9--10：法律理解明显较强，术语和问题使用准确，能主动围绕法律争点展开。

\textbf{信息披露行为。} 0--2：明显回避、隐瞒或拒绝提供关键事实。3--4：披露不充分，对不利信息回避较多，需要反复追问。5--6：能提供主要信息，但完整性一般，偶有遗漏或保留。7--8：总体愿意配合，能较主动提供关键事实和证据线索。9--10：非常坦诚且完整，主动补充关键背景、细节和不利信息。

\textbf{情绪稳定性表现。} 0--2：明显失控、强烈对抗或严重影响沟通。3--4：情绪波动较大，抱怨、指责或偏题较多。5--6：有一定情绪，但整体还能继续围绕案件交流。7--8：整体较冷静，偶有情绪表达但能很快回到正题。9--10：全程稳定、理性、克制，沟通高度聚焦案件事实和需求。

\textbf{叙事组织能力。} 0--2：叙述非常混乱，大量跳跃、重复或矛盾，难以理解。3--4：结构较差，时间线和重点不清，需要频繁澄清。5--6：基本能说清主要经过，但仍有一定跳跃或冗余。7--8：叙事较有条理，时间线和关键节点基本清楚。9--10：叙事清晰完整，结构严谨，重点突出，便于直接提炼案件事实。

\textbf{输入与输出。} 输入为【对话记录】：\texttt{\{dialogue\}}。输出 JSON 键为 \texttt{legal\_literacy}、\texttt{information\_disclosure}、\texttt{emotional\_stability} 和 \texttt{narrative\_proficiency}；每项包含整数 \texttt{score} 和简短 \texttt{evidence}。
\end{CJK*}
\end{promptfigure}

\begin{promptfigure}{Prompt of Persona-Validation LLM-as-Judge (English Version)}{Prompt of Persona-Validation LLM-as-Judge (\textit{English Version})}
You are an expert evaluator of legal-scenario dialogue. Based only on the chat record below, score the party's observed behavior on four dimensions. The evaluated subject is the party, not the lawyer.

\textbf{Requirements.} 1. Score only from the chat record. Do not infer or introduce persona settings, labels, experimental groups, or background information that does not appear in the dialogue. 2. Use integer scores from 0 to 10; higher scores mean stronger observed performance on the dimension. 3. Score bands are: 0--2 very poor, 3--4 poor, 5--6 average, 7--8 good, and 9--10 excellent. 4. If evidence for a dimension is insufficient, score conservatively based on visible behavior; do not default to a high score. 5. The \texttt{evidence} field only needs a brief summary of one or two concrete behaviors in the chat record. 6. Output strict JSON only, with no code block, explanation, or extra text.

\textbf{Legal literacy.} 0--2: almost no understanding of legal relationships or procedure, confused wording, and inability to express legal issues. 3--4: a small amount of legal expression but clearly insufficient understanding, often confusing concepts or procedure. 5--6: basic understanding of core legal issues and can follow the lawyer's discussion, but expression remains rough. 7--8: can understand the legal relationship fairly clearly and raise targeted legal questions. 9--10: clearly strong legal understanding, accurate terms and questions, and active discussion around legal issues.

\textbf{Information disclosure.} 0--2: clearly avoids, hides, or refuses to provide key facts. 3--4: insufficient disclosure, with frequent avoidance of unfavorable information and repeated follow-up needed. 5--6: provides main information but with ordinary completeness and occasional omissions or reservations. 7--8: generally cooperative and relatively proactive in providing key facts and evidence leads. 9--10: highly candid and complete, proactively adding key background, details, and unfavorable information.

\textbf{Emotional stability.} 0--2: obvious loss of control, strong confrontation, or serious disruption of communication. 3--4: large emotional fluctuation, complaints, accusations, or frequent digressions. 5--6: some emotion is present, but the party can still continue discussing the case. 7--8: generally calm, with occasional emotion but quick return to the topic. 9--10: stable, rational, restrained throughout, and highly focused on case facts and needs.

\textbf{Narrative proficiency.} 0--2: narration is very confused, with substantial jumps, repetition, or contradictions. 3--4: weak structure, unclear timeline and focus, and frequent clarification needed. 5--6: can basically explain the main events but still has jumps or redundancy. 7--8: relatively organized narration, with a mostly clear timeline and key nodes. 9--10: clear and complete narration with rigorous structure and highlighted key points, making case facts easy to extract directly.

\textbf{Input and output.} The input is the dialogue record: \texttt{\{dialogue\}}. Return JSON keys \texttt{legal\_literacy}, \texttt{information\_disclosure}, \texttt{emotional\_stability}, and \texttt{narrative\_proficiency}; each item contains integer \texttt{score} and brief \texttt{evidence}.
\end{promptfigure}

\begin{promptfigure}{Prompt of Experimental LLM-as-Judge Evaluation (Chinese Version)}{Prompt of Experimental LLM-as-Judge Evaluation (\textit{Chinese Version})}
\begin{CJK*}{UTF8}{gbsn}
\textbf{Shared system prompt.}
你是一名严谨、客观的法律仿真系统评估专家，具备中国民事诉讼法专业背景。你的任务是依据下文给出的评分维度与标准对仿真系统输出进行 10 分满分评分。你必须严格遵守：仅依据评分标准给出分数，不引入额外标准；每个维度都必须给出简短理由，理由须引用待评材料中的具体内容；不同维度必须独立评分，先逐维度写理由，再给分数；输出严格遵循 JSON 格式，禁止输出任何 JSON 以外的文本。

\textbf{Exp. 1.1 judicial-output alignment.}
任务：对仿真裁判文书与真实裁判文书进行三个独立维度的对齐评分。阶段为 FIT（民事一审）或 SIT（民事二审）。输入为真实裁判文书原文和仿真裁判文书原文。维度 1 是诉求支持对齐：分别提取判决主文，逐项对比原告/上诉人各项诉讼请求的判决结果，优先看核心诉求项的支持、驳回、部分支持方向是否一致。维度 2 是裁判理由（说理）对齐：定位“本院认为”段落，对比核心争点识别、责任分配、主要责任判断和结论通向路径。维度 3 是法律适用准确性：提取援引法条并判断核心法律规则是否对应，不因辅助法条数量、排序或条号细节机械降分。输出 JSON 键分别为 \texttt{claim\_alignment}、\texttt{reasoning\_alignment} 和 \texttt{legal\_citation\_alignment}；每项包含整数 \texttt{score} 和 \texttt{reasoning}。

\textbf{Exp. 1.2 stage authenticity.}
任务：对仿真系统在指定阶段的完整对话/产出进行流程真实性评分。本实验只评程序合规性和流程衔接合理性，不评文书本体内容、实体判断、法律分析、法院名称、诉讼费、日期案号等填写细节；关注阶段错位、程序顺序错误、关键环节缺失或流程断裂。每个维度采用 0--10 分整数制，可以使用 0 到 10 中任意整数，不只使用 2/4/6/8/10 档位。输入包括阶段名称、案件背景和该阶段完整对话/产出。输出 JSON 键为 \texttt{procedural\_compliance} 与 \texttt{process\_coherence}。

程序合规性：9--10 分表示流程完整或几乎完整复刻真实民诉程序，关键环节齐全；7--8 分表示流程基本规范，仅有轻微缺漏；5--6 分表示存在 2--3 处程序问题或环节缺失；3--4 分表示主要程序环节缺失或顺序错误；0--2 分表示程序混乱，明显违反民诉法基本规则。流程衔接合理性：9--10 分表示阶段内推进自然流畅，轮次衔接合理；7--8 分表示整体连贯，个别衔接突兀；5--6 分表示多次衔接不自然，存在跳步或循环；3--4 分表示频繁出现逻辑断裂或重复轮次；0--2 分表示阶段推进混乱，无法形成连贯流程。

\textbf{Exp. 1.3 role consistency.}
任务：对指定 Agent 在仿真案件中的发言进行角色一致性评分。输入包括阶段名称、Agent 角色、案件背景和该 Agent 在本阶段的发言。两个维度独立评分，均使用 0--10 任意整数。输出 JSON 键为 \texttt{stance\_authenticity} 与 \texttt{role\_distinguishability}。

立场与动机真实性：9--10 分表示动机立场始终或几乎始终自洽，符合角色身份；7--8 分表示立场基本一致，偶有不自然之处；5--6 分表示偶有动机偏离，但未根本错位；3--4 分表示多次出现立场不一致或动机模糊；0--2 分表示动机严重错位，例如原告律师替被告辩护。角色间区分度：9--10 分表示高度可区分，角色特征鲜明；7--8 分表示基本可区分，特征较为明显；5--6 分表示部分可区分，特征不够鲜明；3--4 分表示风格趋同，难以与其他角色区分；0--2 分表示风格混淆，无法识别角色身份。
\end{CJK*}
\end{promptfigure}

\begin{promptfigure}{Prompt of Experimental LLM-as-Judge Evaluation (English Version)}{Prompt of Experimental LLM-as-Judge Evaluation (\textit{English Version})}
\textbf{Shared system prompt.}
You are a rigorous and objective evaluator of legal simulation systems with expertise in Chinese civil procedure. Your task is to score the simulation output on a 10-point scale according to the dimensions and criteria below. You must strictly follow these rules: score only according to the given criteria and introduce no extra criteria; give a concise reason for every dimension, citing specific content from the material under evaluation; score different dimensions independently, writing reasons dimension by dimension before assigning scores; output strictly in JSON format and output no text outside JSON.

\textbf{Exp. 1.1 judicial-output alignment.}
Task: score the alignment between the simulated judgment and the real judgment on three independent dimensions. The stage is FIT for first-instance civil judgment or SIT for second-instance civil judgment. The input consists of the real judgment text and the simulated judgment text. Dimension 1 is claim-support alignment: extract the dispositive parts of both judgments and compare, item by item, whether the outcomes for each plaintiff or appellant claim are supported, rejected, or partially supported in the same direction, with priority on core claims. Dimension 2 is adjudicative-reasoning alignment: locate the reasoning section and compare the identification of core issues, allocation of responsibility, main liability judgment, and path from reasons to conclusion. Dimension 3 is legal-application accuracy: extract cited legal provisions and judge whether the core legal rules correspond, without mechanically reducing the score because of auxiliary provision count, order, or article-number details. Output JSON keys are \texttt{claim\_alignment}, \texttt{reasoning\_alignment}, and \texttt{legal\_citation\_alignment}; each contains integer \texttt{score} and \texttt{reasoning}.

\textbf{Exp. 1.2 stage authenticity.}
Task: score the procedural authenticity of the simulation system's complete dialogue or output in the specified stage. This experiment only evaluates procedural compliance and reasonable process continuity; it does not evaluate document content, substantive judgment, legal analysis, court names, litigation fees, dates, case numbers, or other filling details. Focus on stage mismatch, wrong procedural order, missing key links, or process breaks. Each dimension uses any integer from 0 to 10, not only 2/4/6/8/10 anchors. The input includes the stage name, case background, and complete dialogue or output for that stage. The output JSON keys are \texttt{procedural\_compliance} and \texttt{process\_coherence}.

Procedural compliance: 9--10 means the process completely or almost completely reproduces real civil procedure and key links are present; 7--8 means the process is basically standard with only minor omissions; 5--6 means there are 2--3 procedural problems or missing links; 3--4 means major procedural links are missing or in the wrong order; 0--2 means the procedure is confused and clearly violates basic civil-procedure rules. Process coherence: 9--10 means the stage advances naturally and smoothly with reasonable turn-to-turn links; 7--8 means the overall process is coherent with occasional abrupt transitions; 5--6 means multiple transitions are unnatural, with skipped steps or loops; 3--4 means frequent logical breaks or repeated turns; 0--2 means the stage advances chaotically and cannot form a coherent process.

\textbf{Exp. 1.3 role consistency.}
Task: score role consistency for the specified agent's utterances in the simulated case. The input includes stage name, agent role, case background, and the agent's utterances in the stage. The two dimensions are scored independently, each using any integer from 0 to 10. The output JSON keys are \texttt{stance\_authenticity} and \texttt{role\_distinguishability}.

Stance and motivation authenticity: 9--10 means the motivation and stance are always or almost always self-consistent and fit the role identity; 7--8 means the stance is basically consistent with only occasional unnatural moments; 5--6 means there are occasional motivational deviations but no fundamental displacement; 3--4 means repeated inconsistent stance or unclear motivation; 0--2 means severe motivational displacement, such as a plaintiff lawyer arguing for the defendant. Role distinguishability: 9--10 means the role is highly distinguishable and role traits are clear; 7--8 means the role is basically distinguishable with visible traits; 5--6 means the role is partly distinguishable but not vivid; 3--4 means style converges and the role is hard to distinguish from others; 0--2 means style is confused and the role identity cannot be identified.
\end{promptfigure}

\end{document}